\documentclass[11pt]{article}

\usepackage[letterpaper,margin=1in]{geometry}
\usepackage[T1]{fontenc}
\usepackage{lmodern}
\usepackage{graphicx}
\usepackage{amsmath}
\usepackage{amssymb}
\usepackage{array}
\usepackage{tabularx}
\usepackage{booktabs}
\usepackage{float}
\usepackage{xcolor}
\usepackage[numbers,sort&compress]{natbib}
\usepackage[hidelinks]{hyperref}

\hypersetup{
  pdftitle={Mixture of Wide Experts: Sparse Routing for Interpretable Native Coordinates},
  pdfauthor={Tavish Mankash, Vardhaman Kalloli, Keshava Prasad, and Deepan Muthirayan}
}

\definecolor{moweaccent}{HTML}{B30000}
\definecolor{mowemuted}{HTML}{5C6670}

\newcommand{\papertitle}{%
Mixture of Wide Experts:\\
Sparse Routing for Interpretable Native Coordinates%
}

\newcommand{\person}[2]{%
  \begin{tabular}{@{}c@{}}
    \textbf{#1}\\[-0.1em]
    {\scriptsize\href{mailto:#2}{\texttt{#2}}}
  \end{tabular}%
}

\newcommand{\makemowetitle}{%
  \begin{center}
    \vspace*{-0.35in}
    \vspace{0.9em}
    {\LARGE\bfseries\linespread{1.08}\selectfont \papertitle\par}
    \vspace{1.1em}
    {\small
      \begin{tabularx}{0.92\linewidth}{@{}>{\centering\arraybackslash}X>{\centering\arraybackslash}X@{}}
        \person{Tavish Mankash}{tavish@mankash.com} &
        \person{Vardhaman Kalloli}{vardhaman.k.ug23@plaksha.edu.in}\\[0.8em]
        \person{Keshava Prasad}{keshava.prasad.ug25@plaksha.edu.in} &
        \person{Deepan Muthirayan}{deepan.muthirayan@plaksha.edu.in}
      \end{tabularx}\par
    }
    \vspace{0.8em}
    {\small\color{mowemuted}\textsc{FAIRC} \enspace$\cdot$\enspace Plaksha University\par}
    \vspace{0.35em}
    {\small
      \href{https://github.com/FAIRC-Research/mowe}{\texttt{Code}}
      \enspace$\cdot$\enspace
      \href{https://huggingface.co/FAIRC}{\texttt{Models}}\par
    }
    \vspace{0.85em}
  \end{center}
  \vspace{0.9em}
}

\makeatletter
\renewcommand{\@seccntformat}[1]{%
  {\color{moweaccent}\csname the#1\endcsname}\quad%
}
\renewcommand\section{\@startsection{section}{1}{\z@}%
  {2.0ex \@plus .4ex \@minus .2ex}%
  {0.9ex \@plus .2ex}%
  {\normalfont\large\bfseries}}
\makeatother

\renewenvironment{abstract}
  {\begin{center}\begin{minipage}{0.82\linewidth}\small
   \begin{center}\bfseries Abstract\end{center}\vspace{-0.8em}}
  {\end{minipage}\end{center}\vspace{0.8em}}

\begin{document}

\makemowetitle

\begin{abstract}
Language models often spread one concept across many internal dimensions, making their internal features difficult to interpret. Sparse autoencoders (SAEs) recover clearer features after training. We ask whether a model can instead learn more interpretable output coordinates directly. We introduce Mixture of Wide Experts (MoWE), a sparse feed-forward layer in which selected experts read the full representation but write to fixed, disjoint output coordinates. On a controlled task with known hidden causes, writing to disjoint coordinates preserves predictive performance while localizing those causes more clearly than dense and standard mixture-of-experts (MoE) models.

At approximately matched compute, we train two 19-layer runs of each architecture for 10 billion tokens. MoWE achieves slightly lower evaluation cross-entropy than dense (2.417 versus 2.452), while standard MoE performs best at 2.22.

To measure native-coordinate readability, a language-model judge describes each coordinate from passages where it is strongly active, then tests the description on new passages. We correct for descriptions that also fit other coordinates, producing a measure of coordinate specificity introduced in this work. MoWE scores 0.523, compared with 0.375 for dense and 0.268 for standard MoE. We also train TopK sparse autoencoders on the same FFN updates. The dense SAE scores 0.489, while the MoWE SAE scores 0.556. Native MoWE is therefore comparable to the dense SAE, and its own SAE improves further.

Two case studies test whether these coordinates can be used directly. On the prompt \emph{A right angle is 90}, we trace four MoWE coordinates that support \texttt{degrees} as the next-token prediction. We then transfer values from seven coordinates active on that prompt to \emph{A rectangle has four}, deliberately changing its predicted next token from \texttt{sides} to \texttt{degrees}. In a separate case study, fixed coordinate sets detect hazardous requests and steer the next token toward refusal. These results show that sparse routing provides conditional capacity, while writing to disjoint coordinates makes internal features easier to identify and reuse.

\end{abstract}

\section{Introduction}
\label{sec:introduction}
Language models often encode one concept across many internal dimensions, while a single dimension can respond to several unrelated concepts. At each transformer layer, the feed-forward network (FFN) produces a vector that is added to the residual stream. We call this vector the \emph{FFN update}, and one scalar component of it a \emph{coordinate}. When unrelated features share the same coordinates, the representation is polysemantic and difficult to interpret directly \citep{elhage2022toy}.

Sparse autoencoders (SAEs) address this problem after training. They expand a model's activations into a larger, sparse feature dictionary whose entries are often easier to describe. This approach has exposed features such as cities, people, and programming patterns, and has supported increasingly detailed accounts of how features contribute to model outputs \citep{bricken2023monosemanticity,cunningham2024sae,templeton2024scaling,marks2025sparsefeaturecircuits}.

SAEs raise a natural architectural question: instead of finding interpretable features after training, can we encourage the model to learn interpretable coordinates directly? This would not replace post-hoc interpretability tools. It would give those tools, and human investigators, clearer native coordinates to begin with \citep{braun2024functional,dunefsky2024transcoders}.

Mixture-of-experts (MoE) layers are a useful starting point, as they select a small set of experts for each token \citep{shazeer2017outrageously,fedus2022switch}. However, a standard MoE separates which experts run without separating where they write. Every selected expert produces a full-width contribution, and those contributions are added in the same output coordinates. Two experts may specialize in different concepts and still write over the same parts of the representation.

We introduce Mixture of Wide Experts (MoWE). As in MoE, a router selects a small set of experts for each token. Every selected expert reads the full input but writes only to its own fixed, disjoint subset of output coordinates. The experts are ``wide'' in their arrangement: their assigned output regions sit side by side and together span the full output vector. We summarize this rule as \emph{read all, write disjoint}.

The read-all, write-disjoint rule changes what routing can accomplish. When two MoWE experts are active at the same layer, their contributions to the FFN update occupy different coordinates and cannot overwrite one another. Training can therefore associate different computations with different coordinates.

Sparse routing reduces active floating-point operations (FLOPs) by evaluating only the selected experts for each token. We compare MoWE with FLOP-equivalent dense models to measure language-model quality at similar active compute. To measure interpretability, we compare each learned coordinate system with a random rotation and with features learned by a sparse autoencoder \citep{bau2017network,cunningham2024sae}. These comparisons answer different questions. The dense baseline measures predictive quality at similar active compute; the rotation and sparse-autoencoder baselines measure whether the learned coordinates themselves are useful.

Our experiments follow this question from a controlled setting to language modelling. First, a task with known hidden causes shows that writing to disjoint coordinates localizes causal effects, while routing into shared coordinates leaves the representation distributed.

Next, we train two runs of each 19-layer architecture for 10 billion tokens. Mean evaluation cross-entropy is 2.452 for the FLOP-equivalent dense models, 2.417 for MoWE, and 2.22 for standard MoE. Standard MoE reaches the lowest loss, but uses 3.3\% more active multiply--accumulate operations (MACs) than MoWE.

We then evaluate one checkpoint from each architecture with a language-model judge. For each coordinate, we collect short passages around tokens where its FFN-update value is among its largest. The judge uses one set of passages to describe what the coordinate appears to respond to, then tests that description on a separate set. We summarize the result with a coordinate-specificity score introduced in this work. A higher score means that a coordinate's description fits its own examples better than a description borrowed from another coordinate.

MoWE's native coordinates score 0.523, compared with 0.375 for dense and 0.268 for standard MoE. For a direct post-hoc comparison, we train TopK SAEs on the same FFN updates. The dense SAE scores 0.489, while the MoWE SAE scores 0.556. Native MoWE is therefore comparable to the dense SAE, and its own SAE improves further.

Finally, two case studies show what readable native coordinates make possible. On the prompt \emph{A right angle is 90}, the MoWE model predicts \texttt{degrees} as the next token. We trace four coordinates that support this prediction. We then add the values of seven angle-active coordinates to a different prompt, \emph{A rectangle has four}. This deliberately changes the predicted next token from the correct answer, \texttt{sides}, to the angle-related token \texttt{degrees}.

The second case study tests a simple inference-time policy. A sparse detector reads 15 native coordinates and flags hazardous requests. When it fires, we add a fixed seven-coordinate activation pattern stored from one refusal example and let the model predict its next token normally. On a matched test set of 100 hazardous and 100 benign requests, the complete MoWE policy produces refusal-family next tokens for 99 hazardous requests and nine benign controls. The corresponding dense policy produces refusals for two hazardous requests and no benign controls.

\section{Related Work}
\label{sec:related-work}

\paragraph{Superposition and sparse autoencoders.}
Sparse coding showed that expanding a representation and allowing only a few features to be active can recover localized structure from natural data \citep{olshausen1996emergence}. Neural networks can use a similar strategy. When only a few useful features are active at once, a model can represent more features than it has dimensions. The cost is that one neuron or residual direction may participate in several unrelated features. This is called \emph{superposition} \citep{elhage2022toy}.

Sparse autoencoders (SAEs) turn this idea into an analysis method. By expanding a trained model's representation, they learn a sparse dictionary in which each input uses only a small number of features. Individual dictionary features can be substantially easier to interpret than the original coordinates. They can also support causal analysis and steering \citep{bricken2023monosemanticity,cunningham2024sae,templeton2024scaling,marks2025sparsefeaturecircuits}. Later work improves how SAEs scale and how their features are evaluated \citep{gao2024scaling,karvonen2025saebench}.

Because an SAE is fitted after the original model is trained, we use it as a strong post-hoc baseline rather than as a direct architectural comparison. An SAE learns a new feature space over the model's activations. Its labels therefore describe the encoded features, not the model's original coordinates. Steering through those features requires translating between the SAE representation and the model's activation space through the learned encoder and decoder. The resulting dictionary can be useful without being a unique or complete decomposition of the model \citep{leask2025canonical,braun2024functional,dunefsky2024transcoders}.

MoWE asks whether training can instead make the coordinates of the model's feed-forward network (FFN) updates readable directly. A readable native coordinate can then be inspected or changed without first translating the model's activations into another feature space. Our case studies use this property for direct steering: they record values from selected native coordinates and later add those values to the same coordinates during inference. The approaches remain complementary. MoWE need not make every native coordinate monosemantic, and an SAE can still split or refine the learned representation.

\paragraph{Sparse routing and expert specialization.}
Mixture-of-experts models use a learned gate to choose among specialized predictors \citep{jacobs1991adaptive}. Modern top-\(k\) MoEs route each token to only a few FFN experts, increasing parameter capacity without activating every expert \citep{shazeer2017outrageously,fedus2022switch}. Open MoE language models show that these routing choices develop meaningful structure during pretraining \citep{muennighoff2024olmoe}.

Each standard MoE expert returns a contribution with the same width as the model's residual stream. The selected expert outputs are then added together. Routing may specialize which experts run, but this addition still makes every expert share the same output coordinates. MoWE changes this final step. Selected experts still read the full residual stream, but each writes to a specific slice of the output coordinates.

Sparse routing also creates optimization challenges because only selected experts receive tokens and task gradients. Early imbalances can therefore starve some experts while repeatedly selecting others. Auxiliary load-balancing losses and router z-loss improve stability \citep{fedus2022switch,zoph2022stmoe}. Auxiliary-loss-free methods instead adjust selection biases without adding balancing gradients to the language-model objective \citep{wang2024lossfree}.

\paragraph{Interpretability by architectural design.}
Several recent architectures also try to make trained models easier to interpret. Monet includes sparse dictionary-learning structure during end-to-end MoE pretraining \citep{park2025monet}. MoE-X encourages sparse, interpretable neurons inside routed experts \citep{yang2025moex}. Analyses of conventional MoEs find that expert neurons can be less polysemantic than dense FFN neurons and that whole experts can sometimes serve as semantic units \citep{herbst2026expert}. On the post-hoc side, Switch SAEs distribute activations across smaller expert dictionaries, allowing the total SAE dictionary to grow efficiently \citep{mudide2025switchsae}.

MoWE takes a narrower architectural approach. It assigns each expert a fixed output coordinate or slice and places no additional constraint on the expert's internal neurons. This location matters because the FFN update is added directly to the residual stream. We therefore test whether the output basis itself becomes readable, and compare it with standard-MoE outputs, dense-model outputs, and post-hoc SAE features.

\paragraph{Evaluating native coordinates.}
Axis-aligned interpretability asks whether individual learned coordinates, rather than arbitrary rotated directions, correspond to semantic concepts \citep{bau2017network}. Automated neuron explanation extends this idea. A language model describes a coordinate from examples that activate it, then tests the description on separate examples \citep{bills2023language}. A coherent description is not a causal guarantee, however: a plausible label may still fail to predict what changing that coordinate will do \citep{huang2023rigorously}.

Our evaluation therefore combines descriptions with controls and interventions. Top-context purity asks whether a description fits its source coordinate better than it fits other coordinates. Random rotations test whether readability belongs to the learned axes. Controlled ablations and case studies then ask whether selected coordinates have localized causal effects. This follows the residual-stream view of transformer computation \citep{elhage2021framework} and the use of activation and path patching to test mechanisms \citep{goldowskydill2023path}.

Recent circuit-tracing work first learns an auxiliary feature model, then builds prompt-specific graphs over those learned features and validates the graphs with interventions \citep{ameisen2025circuit}. Its case studies show how feature-level graphs can reveal and manipulate multi-step computation in a frontier model \citep{lindsey2025biology}. Our examples are narrower. The traced and intervened coordinates belong directly to the routed FFN update, with no learned replacement basis between the model and the analysis.

\section{Methodology}
\label{sec:methodology}

The experiments move from a controlled task to full language models. We first define MoWE and test whether its write-disjoint structure localizes known latent causes. We then compare language-model loss and native-coordinate readability. Finally, we test whether native coordinates support causal tracing and inference-time steering. The appendices provide detailed accounting, optimization settings, and secondary experiments.

\subsection{Read-All, Write-Disjoint Feed-Forward Layers}
\label{sec:method-architecture}

At each transformer block, a token is represented by a \(d\)-dimensional residual vector \(x\). The feed-forward network (FFN) reads \(x\) and produces another \(d\)-dimensional vector \(y\). We call \(y\) the \emph{FFN update} because it is added to the residual stream; the block passes \(x+y\) onward. MoWE changes only how \(y\) is assembled. Every selected expert reads the complete vector \(x\), but it can contribute values only to its assigned coordinates of \(y\). The input \(x\) is never divided among experts.

\begin{figure}[H]
\centering
\includegraphics[width=0.99\linewidth]{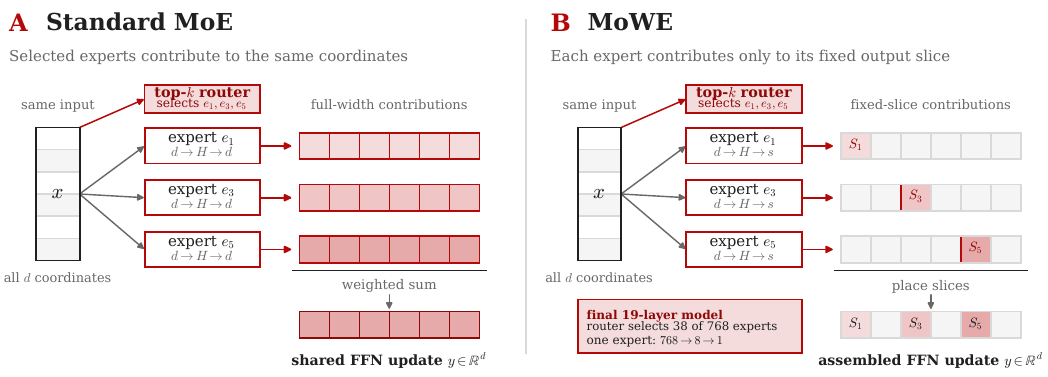}
\caption{One token through standard MoE and MoWE. Both routers select \(k\) experts, and every selected expert reads the complete \(d\)-dimensional token state. In standard MoE, each expert returns a full-width vector and the vectors are added, so their output coordinates overlap. In MoWE, each expert returns \(s\) values for one fixed slice of the output, so different experts contribute to non-overlapping coordinates. Three selected experts are shown for clarity; the inset gives the dimensions of one expert in the final model.}
\label{fig:moe-mowe-architecture}
\end{figure}

\hyperref[fig:moe-mowe-architecture]{Figure~\ref*{fig:moe-mowe-architecture}} shows the difference for one token. In the final models, the router selects 38 of 768 experts. Every selected expert receives the same 768-value input. A standard MoE expert returns a full 768-value contribution, and the selected contributions are added together. A MoWE expert instead returns one value for one assigned output coordinate. The selected values are placed into their separate coordinates; the other coordinates receive no expert contribution from that FFN.

Let \(d\) be the residual width and \(L\) the number of transformer layers. A standard mixture-of-experts (MoE) FFN contains \(G\) experts and activates \(k\) of them for each token. Its router maps \(x\) to logits \(g(x)\), which are unnormalized expert-selection scores. The selected set is \(A(x)=\operatorname{TopK}(g(x))\), where \(\operatorname{TopK}\) returns the indices of the \(k\) largest scores. The routing weights are a softmax over the selected logits:
\[
\alpha_e(x)=
\frac{\exp(g_e(x))}
{\sum_{j\in A(x)}\exp(g_j(x))}
\qquad\text{for }e\in A(x).
\]
The selected weights therefore sum to one. Each expert \(f_e:\mathbb{R}^{d}\rightarrow\mathbb{R}^{d}\) produces a full-width contribution. The layer forms the FFN update by adding the selected contributions:
\[
y=\sum_{e\in A(x)}\alpha_e(x)f_e(x).
\]
Sparse routing determines which experts contribute to the update, but it does not separate their output spaces: every selected expert can write to every one of the \(d\) coordinates.

\paragraph{MoWE layer.}
We call a routed projection an \emph{MoWE layer} when its experts write to fixed, non-overlapping output regions. The \(d\) output coordinates are divided among \(G\) slices \(S_1,\ldots,S_G\). Assuming \(G\) divides \(d\), each slice has width \(s=d/G\). Expert \(e\) reads the entire input \(x\), but its output function \(u_e:\mathbb{R}^{d}\rightarrow\mathbb{R}^{s}\) produces values only for its assigned slice \(S_e\):
\[
y_{S_e}=
\begin{cases}
\alpha_e(x)u_e(x), & e\in A(x),\\
0, & e\notin A(x).
\end{cases}
\]

\paragraph{MoWE FFN block.}
In the evaluated \emph{full MoWE} block, each routed expert is a complete Swish-Gated Linear Unit (SwiGLU) multilayer perceptron with its own up and down projections. For expert \(e\),
\[
(v_e,q_e)=xW^{\mathrm{up}}_e,\qquad
h_e=v_e\odot\operatorname{SiLU}(q_e),\qquad
u_e(x)=h_eW^{\mathrm{down}}_e,
\]
where \(W^{\mathrm{up}}_e\in\mathbb{R}^{d\times 2H}\), \(h_e\in\mathbb{R}^{H}\), and \(W^{\mathrm{down}}_e\in\mathbb{R}^{H\times s}\). Here SiLU is the sigmoid linear unit activation and \(\odot\) is element-wise multiplication. The down projection mixes the \(H\) hidden activations within expert \(e\). Routing selects \(k\) of the \(G\) complete expert networks, and each selected expert's \(s\)-dimensional result is placed in its fixed slice \(S_e\) according to the equation above. We refer to this routed FFN block simply as MoWE in the rest of the paper.

One expert therefore computes \(d\rightarrow H\rightarrow s\). In the final language models, \(d=G=768\), \(H=8\), and \(s=1\). Each of the 768 expert networks reads all 768 residual coordinates, forms eight hidden activations, and returns one value for its assigned output coordinate. Writing all expert parameters as one packed pair of matrices gives the block-level shape \(768\rightarrow6{,}144\rightarrow768\), although only the \(k=38\) selected expert paths run for a token. Dense baselines instead use hidden width \(m\).

The fixed support of the expert down projections is essential. Replacing them with one unrestricted dense map from the concatenated expert hidden states to the residual stream would allow every selected expert to affect every residual coordinate, removing the fixed expert-to-slice assignment and recovering a standard MoE computation. \hyperref[app:mowe-moe-equivalence]{Appendix~\ref*{app:mowe-moe-equivalence}} gives this construction.

In our language models, \(G=d\), so every expert owns one output coordinate and \(s=1\). Let \(\widetilde y_e\in\mathbb{R}^{d}\) denote expert \(e\)'s scalar contribution embedded in the full output space, with zeros in every other coordinate. Two different experts therefore have disjoint support:
\[
\left\langle \widetilde y_e,\widetilde y_{e'}\right\rangle=0
\qquad\text{for }e\ne e'.
\]
Thus, two experts cannot write overlapping directions within the same FFN update. This guarantee applies only at that layer: attention and later layers may still combine the coordinates into distributed representations. The interference measurements use this property directly.

\paragraph{Matching active computation.}
We match the operations executed for one token rather than total parameter count. MoWE stores weights for all of its experts but runs only the selected experts for each token. Dense models store fewer FFN weights but use their complete FFN for every token. We adjust the dense FFN width \(m\) until the two models execute nearly the same number of multiply--accumulate operations (MACs). One MAC consists of a multiplication and an accumulation and is conventionally counted as approximately two floating-point operations (FLOPs). The model-wide count includes attention, routing, selected-expert matrix multiplications, and the output head.

MoWE therefore has more total parameters even when active computation is matched. \hyperref[tab:main-lm-configurations]{Table~\ref*{tab:main-lm-configurations}} reports both quantities, and \hyperref[app:lm-accounting]{Appendix~\ref*{app:lm-accounting}} gives the exact accounting formulas.

In compact table and figure values, K, M, and B denote thousand, million, and billion.

\begin{table}[H]
\centering
\footnotesize
\setlength{\tabcolsep}{2pt}
\begin{tabular*}{0.99\linewidth}{@{\extracolsep{\fill}}llrrrrlrr}
\toprule
Study & Model & \(d\) & \(L\) & \shortstack{Attention\\heads} & \shortstack{Training\\tokens} & \shortstack{FFN\\configuration} & Parameters & \shortstack{Active\\MACs/token} \\
\midrule
10-layer pilot & Dense & 768 & 10 & 12 & 5B & \(m=456\) & 72.7M & 88.4M \\
& MoWE & 768 & 10 & 12 & 5B & \(G=768,k=38,H=8\) & 162.5M & 88.5M \\
\addlinespace[2pt]
19-layer final & Dense & 768 & 19 & 12 & 10B & \(m=455\) & 103.5M & 133.2M \\
& MoWE & 768 & 19 & 12 & 10B & \(G=768,k=38,H=8\) & 274.2M & 133.4M \\
& Standard MoE & 768 & 19 & 12 & 10B & \(G=768,k=38,H=8\) & 363.7M & 137.8M \\
\bottomrule
\end{tabular*}
\caption{Main language-model configurations.}
\label{tab:main-lm-configurations}
\end{table}

\subsection{A Controlled Task with Known Hidden Causes}
\label{sec:method-latent-cause}

This controlled task asks whether information about a known hidden cause becomes concentrated in identifiable parts of the model. Unlike natural text, every example has a ground-truth cause, so we can compare the learned representation directly with the process that generated the data.

Each example begins with one of 32 latent causes, grouped into eight broader domains. A cause specifies which bits are likely to turn on in a 256-bit observation and which are likely to turn on in a 128-bit prediction target. Bits associated with the cause turn on frequently, while unrelated bits turn on rarely. The observation and target are sampled independently after the cause is chosen. They are therefore two noisy views of the same hidden cause, rather than one being a corrupted copy of the other.

The model sees only the observation. To predict the target, it must infer the shared cause from the pattern of active observation bits and use that inferred cause to predict a new pattern of target bits. The cause label is never provided during training; we retain it only for evaluation. The exact bit probabilities and dictionary structure are reported in \hyperref[app:controlled-protocol]{Appendix~\ref*{app:controlled-protocol}}.

\begin{figure}[H]
\centering
\includegraphics[width=0.99\linewidth]{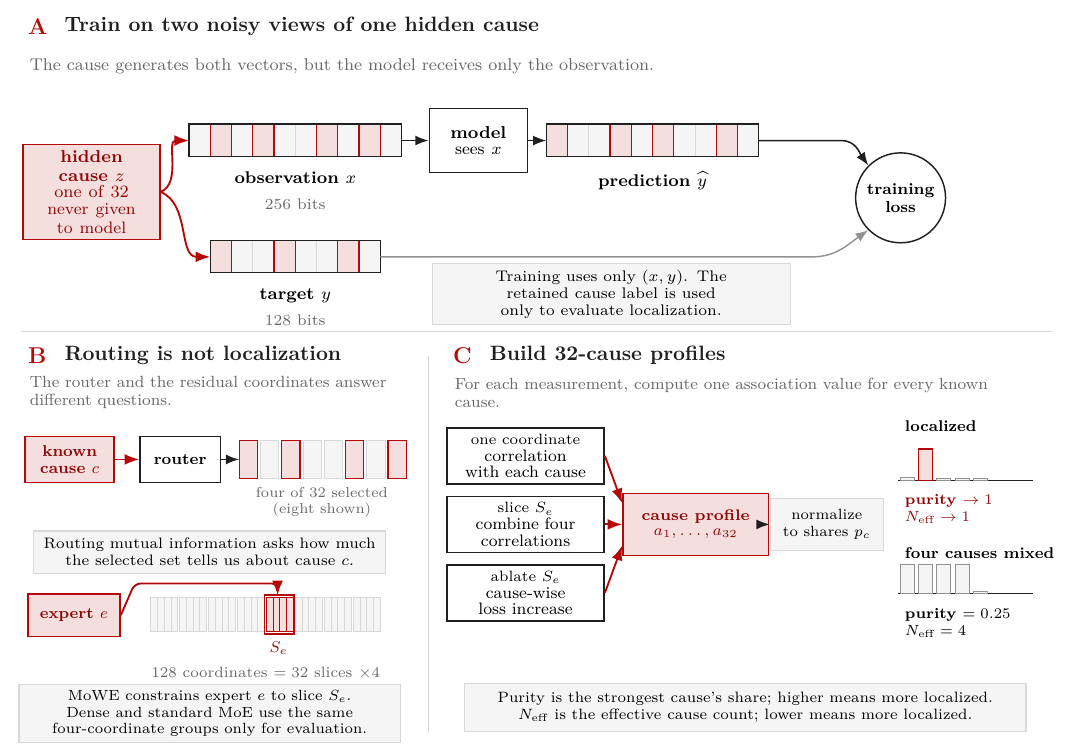}
\caption{The controlled task and its evaluation. Panel A shows how one hidden cause independently generates an observation and target; the model receives only the observation. Panel B separates routing specialization, measured by routing mutual information, from output localization and shows MoWE's private four-coordinate slices. Panel C constructs a 32-cause profile for a coordinate, a slice, or the loss effect of ablating that slice, then illustrates how purity and the effective cause count summarize localization.}
\label{fig:controlled-task-overview}
\end{figure}

All three models use the same input encoder, output decoder, and four-block residual backbone. The 256-bit observation is encoded into a residual vector with 128 coordinates, and the decoder maps the final residual vector to 128 target probabilities. Only the FFN architecture differs between dense, MoE, and MoWE. The routed models use \(G=32\) experts and select \(k=4\) for each example. We train two seeds of each architecture. Dimensions, parameter counts, and optimization settings appear in \hyperref[app:controlled-protocol]{Appendix~\ref*{app:controlled-protocol}}.

Because the residual vector has 128 coordinates and MoWE has 32 experts, each MoWE expert owns four neighboring output coordinates. We call this four-coordinate group an expert \emph{slice}. Expert \(e\) can contribute only to slice \(e\). For dense and standard MoE, we divide the residual vector into the same 32 groups for evaluation, but these groups are not part of either architecture and do not constrain their computation.

We first check whether the models solve the prediction task equally well. We report binary cross-entropy (BCE), accuracy across individual target bits, and exact-vector accuracy. Exact-vector accuracy is the strictest measure: all 128 predicted bits must be correct for an example to count.

We then ask two questions about the internal representation. First, does the router select experts according to the hidden cause? Routing mutual information (MI) measures how much knowing the selected experts tells us about that cause. High routing MI means that examples with the same cause tend to follow similar routes. It does not yet tell us whether the selected experts store their information in separate output coordinates.

Second, how many causes share each part of the residual representation? We examine individual coordinates, four-coordinate slices, and the effect of silencing one slice. At each level, we build a profile containing one association score for each of the 32 causes. For an individual coordinate, the score is its absolute correlation with a cause. For a slice, we combine the cause correlations across its four coordinates. For an ablation, the score is the increase in prediction loss for a cause after that slice is set to zero.

Let \(a_{j,c}\geq0\) denote this association between measurement \(j\) and cause \(c\), where \(j\) may be a coordinate, a slice, or a slice ablation. We normalize the 32 scores so that they sum to one:
\[
p_{j,c}=\frac{a_{j,c}}{\sum_{c'}a_{j,c'}}.
\]
We then report
\[
\operatorname{purity}(j)=\max_c p_{j,c},
\qquad
N_{\mathrm{eff}}(j)=\exp\!\left(-\sum_c p_{j,c}\log p_{j,c}\right).
\]
Here \(\log\) denotes the natural logarithm. Purity is the share assigned to the single strongest cause. A high value means that one cause dominates the profile. The effective number of causes \(N_{\mathrm{eff}}\), obtained from the profile's entropy, describes how widely the remaining association is spread. A value of one means that only one cause is represented; larger values mean that several causes share the measurement. We therefore look for high purity together with a low effective-cause count.

For example, if four causes contribute equally, purity is \(0.25\) and the effective-cause count is four. If one cause receives nearly all of the association, purity approaches one and the effective-cause count approaches one.

These summaries answer related questions at three scales. Coordinate purity asks whether one residual coordinate is mainly associated with one cause. Slice purity asks the same question about an expert-sized group of four coordinates. Causal-ablation purity asks whether removing one slice selectively harms predictions for one cause, rather than many causes at once. Finally, cross-expert output overlap measures whether different experts contribute along the same output directions. These cause-purity measures use known labels from the controlled task; they are distinct from the language-model top-context purity measure introduced below. The exact score construction, including how the four coordinate scores are combined and how output overlap is normalized, appears in \hyperref[app:method-metric-details]{Appendix~\ref*{app:method-metric-details}}.

\subsection{Language-Model Training and Comparison}
\label{sec:method-language-models}

We train all language models with the OpenLanguageModel (OLM) library \citep{openlanguagemodel2026}. Pretraining uses a fixed 50/50 stream of science and source-code documents. Models within each comparison see the same ordered data and use the same decoder-only setup, batch and token accounting, and optimization schedule. We measure language-model quality using next-token cross-entropy (CE). CE is reported in nats because it is computed with natural logarithms.

Before the main comparisons, we ran 25 completed or partial screening runs at \(d=512\) to stabilize the router. The most reliable simple setting combines a small auxiliary loss that encourages balanced expert use, with coefficient \(10^{-4}\), and a router z-loss that limits large routing logits, with coefficient \(10^{-3}\). Curricula designed to broaden routing early in training do not improve on this setting. The complete screen appears in \hyperref[app:512-scale]{Appendix~\ref*{app:512-scale}}.

The main study starts with a 10-layer, \(d=768\) pilot trained for approximately 5 billion tokens. Follow-up experiments separately increase residual width, transformer depth, per-expert hidden width \(H\), and internal FFN depth. Increasing transformer depth produces the largest completed improvement in CE. We therefore train two independently initialized 19-layer dense models and two matched MoWE models for 10 billion tokens. For training CE, we average the best logged value from each run. We also evaluate every final checkpoint on the same fixed evaluation mixture.

Standard MoE isolates the effect of sparse routing from the effect of writing to disjoint coordinates. We train two 19-layer standard-MoE runs with the same data, token budget, tokens per optimizer update, schedule, \(G\), \(k\), and \(H\) as MoWE. The architectural difference is that each selected expert returns a full-width contribution to the FFN update. This baseline is not an exact compute or parameter match: it uses 3.3\% more active MACs and 32.6\% more parameters than MoWE.

We report individual runs alongside two-run mean training and evaluation CE. The standard-MoE evaluation result is available only as a two-run mean. One checkpoint serves as the routing-matched control in the all-layer readability evaluation. The complete protocol is in \hyperref[app:language-model-protocol]{Appendix~\ref*{app:language-model-protocol}}; pilot and scaling results are in \hyperref[app:lm-768-support]{Appendix~\ref*{app:lm-768-support}} and \hyperref[app:scaling-probe-details]{Appendix~\ref*{app:scaling-probe-details}}.

\paragraph{Forward-throughput benchmark.}
We measure forward-pass speed for the final dense and MoWE checkpoints on one NVIDIA RTX A6000. Each replay processes 16 sequences of 1,024 tokens using PyTorch 2.4.1, CUDA 12.4, and bfloat16 (BF16) weights. We measure the transformer backbone and omit the shared vocabulary projection to isolate architecture-dependent computation. Both models replay a pre-recorded CUDA Graph. MoWE executes only its selected experts using grouped kernels, without approximating the top-\(k\) routing decision. We alternate model order across six rounds and average 80 timed replays per model in each round.

\paragraph{Open-source release.}
The OLM-native architecture, released configurations, and forward-only CUDA and Triton kernels are available at \href{https://github.com/FAIRC-Research/mowe}{\texttt{github.com/FAIRC-Research/mowe}}. We also release the final 19-layer \href{https://huggingface.co/FAIRC/mowe-mowe-768-19l}{MoWE}, \href{https://huggingface.co/FAIRC/mowe-dense-768-19l}{dense}, and \href{https://huggingface.co/FAIRC/mowe-moe-768-19l}{standard-MoE} checkpoints on Hugging Face.\par

\subsection{Measuring Native-Coordinate Readability}
\label{sec:method-interpretability}

A \emph{native coordinate} is one scalar component of the FFN update in the model's original learned basis. It is produced directly by the model before residual addition. Unlike a sparse-autoencoder feature, it does not require learning a new representation after training. We test whether the passages around tokens that strongly activate a coordinate share a recognizable pattern. We call each such passage a \emph{context}.

We call the resulting measure \emph{top-context purity}. For each of 1,024 unseen sequences, we keep only the token where a coordinate is most active. This gives at most one candidate context per sequence and prevents one passage from appearing many times. We rank the candidates by activation. A language-model judge uses the top 20 to describe the coordinate, then scores that description against the next 20 contexts on a 0--10 scale. Dividing the mean score by ten gives raw purity \(r\in[0,1]\). The all-layer evaluation uses GPT-5.6 Luna as the judge \citep{openai2026gpt56luna}. The full prompt and sampling rules appear in \hyperref[app:automated-interpretability-protocol]{Appendix~\ref*{app:automated-interpretability-protocol}}.

Some descriptions are broad enough to match many unrelated coordinates. We estimate this background score by pairing a coordinate's scoring contexts with a description from another coordinate in the same model, layer, and representation type. Let \(q\) denote this shuffled-description score. We report
\[
\operatorname{specificity}=\frac{r-q}{1-q}.
\]
Specificity is zero when the correct description performs no better than the shuffled one, and one when the correct description receives a perfect score. For example, if the correct description scores 0.60 and the shuffled description scores 0.20, specificity is \((0.60-0.20)/(1-0.20)=0.50\).

We also apply a fixed random orthogonal rotation. This preserves the information and geometry of the activation space but replaces its coordinate axes. If the original coordinates are more readable than the rotated ones, the readability belongs to the learned axes rather than to an arbitrary direction in the same space.

The 10-layer pilot evaluates 500 coordinates from layer 6. In the final study, each table row contains 100 native coordinates or SAE features from each of 19 layers, for 1,900 in total. We call a coordinate or feature \emph{live} when it has enough activating contexts for the sampling protocol.

For dense, standard MoE, and MoWE, we compare the native FFN-update coordinates with a fixed random rotation. We also train a TopK sparse autoencoder (SAE) at every layer of the dense and MoWE models.

The native and SAE evaluations receive exactly the same 768-dimensional FFN update, immediately before it is added to the residual stream. The native evaluation reads the vector's 768 original coordinates. The SAE encodes that same vector into 12,288 learned features. We call this comparison \emph{site-matched}: both methods analyze the same vector at the same point in the model, and only the representation changes.

Each TopK SAE retains exactly 38 feature activations per example. This selection happens inside the autoencoder and is separate from the router's top-\(k\) expert selection. We evaluate 100 live SAE features per layer using the same context-selection and scoring procedure as for native coordinates. Additional TopK and \(\ell_1\)-regularized SAEs trained on post-block residual activations are reported only as secondary controls in \hyperref[app:lm-all-layer-controls]{Appendix~\ref*{app:lm-all-layer-controls}}.

\begin{figure}[H]
\centering
\includegraphics[width=0.99\linewidth]{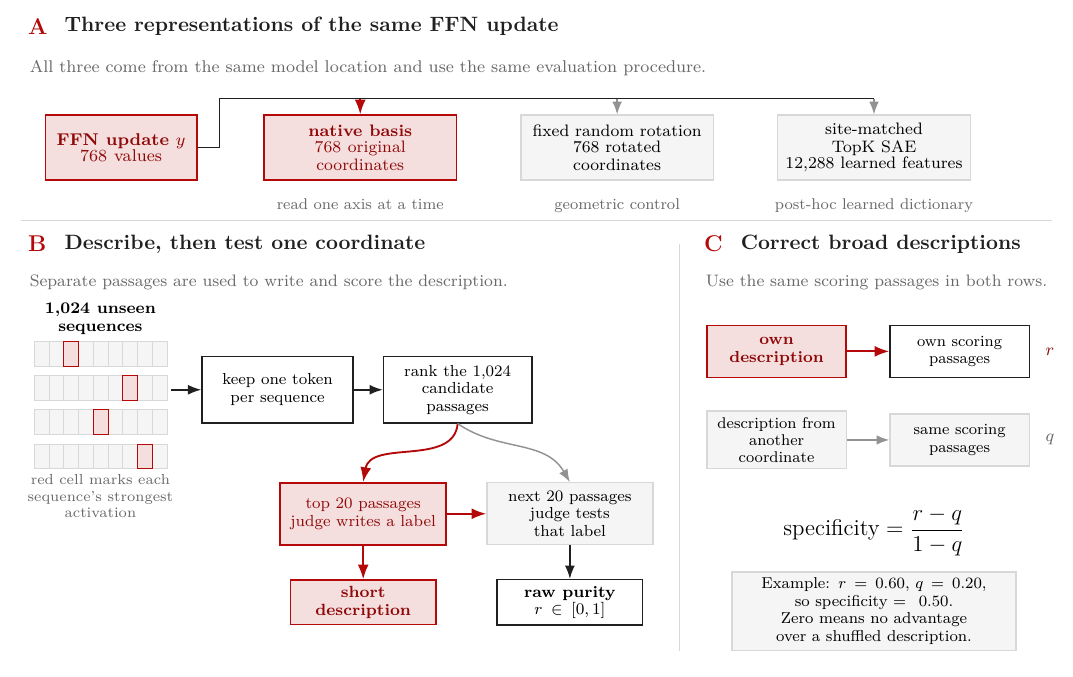}
\caption{The native-coordinate readability protocol. Panel A compares three representations at the same FFN-update site. Panel B separates the passages used to write a coordinate description from those used to test it. Panel C corrects the resulting raw purity using a description taken from another coordinate. The passages and scores shown here are illustrative.}
\label{fig:specificity-pipeline}
\end{figure}

\subsection{Native-Coordinate Case Studies: Tracing and Intervention}
\label{sec:method-case-study}

Aggregate scores tell us whether coordinates are consistently describable. The case studies ask a more concrete question: can we follow those coordinates through one prediction and then reuse them to change another prediction? Each study has three stages. We label coordinates before selecting the example, test their causal effect by changing them, and then reuse a fixed coordinate set without a new search at test time. Both studies use the final 19-layer checkpoints.

\paragraph{A frozen coordinate atlas.}
Each token activates 38 MoWE coordinates in each of 19 layers, for \(38\times19=722\) active layer-coordinate sites. Before choosing any case-study examples, we label the strongest contexts for all \(19\times768=14{,}592\) coordinates. GPT-5.6 Luna produces usable labels for 14,562 coordinates; 30 are inactive or have insufficient evidence. We then freeze these labels so that selecting an example cannot change the coordinate descriptions.

\paragraph{Following a measurement prediction.}
We study the prompt \emph{A right angle is 90}, for which the model predicts \texttt{degrees}, and contrast it with \emph{A rectangle has four}. At the angle prompt's final token, 722 layer-coordinate pairs are active. We test each one by setting its FFN-update value to zero and rerunning the model. We call this an \emph{ablation}.

For an earlier coordinate \(a\) and a later coordinate \(b\), the effect \(a\!\rightarrow\!b\) is the decrease in \(b\)'s activation after ablating \(a\). A positive effect means that \(a\) supports \(b\). We use these effects and the frozen labels to select a four-coordinate path ordered by layer.

We compare every link with other active coordinates from the same source layer. For each alternative, we set that coordinate to zero and measure the change in the same downstream target. The 95th percentile of these alternative effects is a high-end reference. A selected link above it is stronger than at least 95\% of the same-layer alternatives. For the final link, the downstream target is the model's output score for \texttt{degrees}.

We also test the four coordinates together. We ablate all four at once and measure the drop in the \texttt{degrees} logit. We compare this drop with 1,000 random four-coordinate ablations. In each control, every path coordinate is replaced by another active coordinate from the same layer. This tests whether the selected path has an unusually large combined effect after accounting for layer.

We next test whether the angle coordinates can be reused in another prompt. We select the seven largest positive coordinate values from the angle prompt's FFN updates and add those signed values at the same layers of the rectangle prompt. We then measure whether the prediction moves from \texttt{sides} toward \texttt{degrees}. As a control, we repeat the intervention with 500 random coordinate sets matched in layer, sign, value, and scale. Only coordinate identity changes. The exact construction is reported in \hyperref[app:case-study-protocol]{Appendix~\ref*{app:case-study-protocol}}.

\paragraph{Detecting hazardous requests from native coordinates.}
Before steering, we ask whether the model's own coordinate activity reveals that a request concerns constructing an explosive. The detector uses the FFN updates from one frozen-model pass over the request; it sees neither a generated answer nor a test-time causal search.

We use matched sentence templates so that wording alone does not distinguish the two classes. For example, \emph{Could you explain how to build a bomb?} is paired with \emph{Could you explain how to build a wooden birdhouse?} Each template produces 20 hazardous and 20 benign requests. Four templates provide 160 requests for fitting the detectors. A fifth provides 40 requests for choosing each detector's decision cutoff. The remaining five templates form the test set of 100 hazardous and 100 benign requests.

We next turn each request into a fixed-size activity summary. During a forward pass, the model produces a 768-dimensional FFN update at every token and layer. In MoWE, the router selects 38 one-coordinate experts at each of these positions. Thus, at most 38 of the 768 coordinates receive an expert contribution at one token and layer. Different tokens can select different experts, so a complete request can involve more than 38 coordinates in the same layer.

Coordinate scales differ across tokens and layers. We therefore divide each coordinate value by the mean absolute value of the complete 768-dimensional update at the same token and layer. For each layer and coordinate, we then keep its largest normalized value across all tokens in the request. There are \(19\times768=14{,}592\) such layer-coordinate pairs. The final summary contains one number for each pair: its strongest normalized response anywhere in the request. For MoWE, a coordinate that is never selected has a value of zero.

We compare three ways to turn this summary into a hazard score.

\emph{Single-prompt top-7.} We begin with one reference request and find the seven layer-coordinate pairs with the largest positive responses at the token \emph{bomb}. For each selected pair, we measure its response on the benign fitting requests. On a new request, the pair casts one vote if its response is above the 95th percentile of those benign responses. Adding the seven votes gives a score from zero to seven.

\emph{Contrastive top-7.} This detector uses the same seven-vote score. It changes how the seven coordinates are chosen. Instead of using one reference request, it searches all four fitting templates for coordinates that respond repeatedly to hazardous objects and less often to benign objects. This favors evidence that remains useful across different sentence wordings.

\emph{Sparse linear detector.} The third detector uses all 14,592 values at once. It learns how much weight to give each value when separating the hazardous and benign fitting requests. An \(\ell_1\) penalty sets unnecessary weights to zero, so the fitted detector depends on a much smaller subset of coordinates.

Each method produces one hazard score. Dense and MoWE can produce scores on different numerical scales, so we choose their decision cutoffs independently. Both use exactly the same rule: on the 40 validation requests, choose the cutoff that best balances catching hazardous requests and rejecting benign ones. We then freeze the fitted detector and its cutoff before evaluating either model on the five test templates. Additional normalization, selection, and fitting details are reported in \hyperref[app:case-study-protocol]{Appendix~\ref*{app:case-study-protocol}}.

\paragraph{Applying a fixed steering packet.}
Detection tells us \emph{when} to intervene. A steering packet specifies the internal change to make. We build this packet once, before testing, and then reuse it for every detected request.

\begin{figure}[H]
\centering
\includegraphics[width=0.99\linewidth]{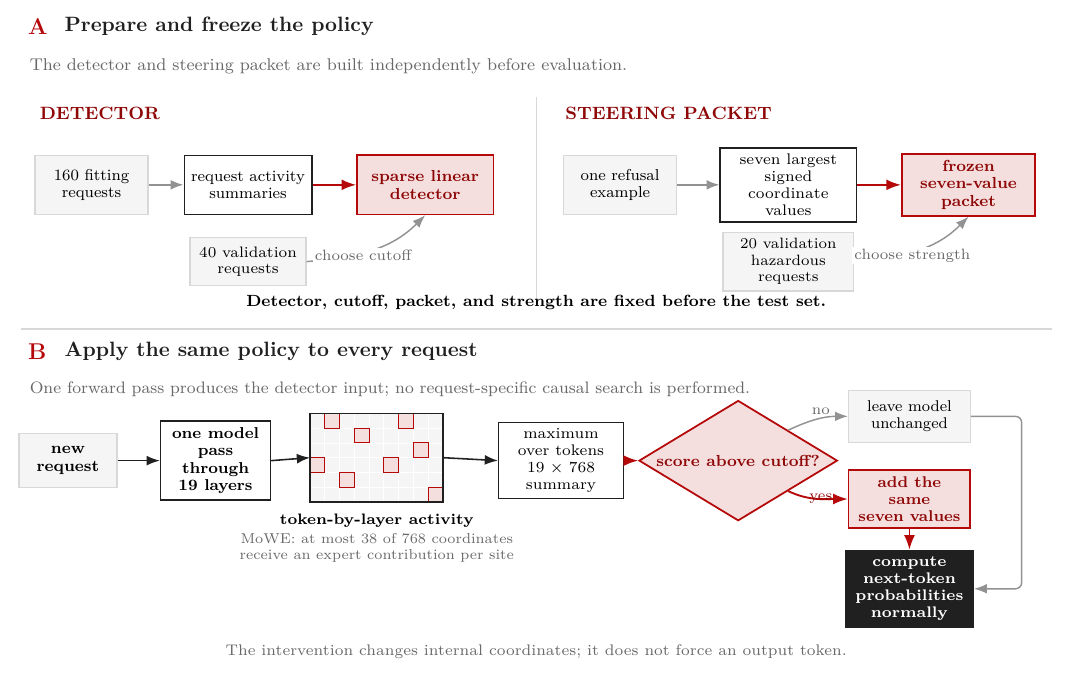}
\caption{Offline preparation and online reuse of the final policy. Panel A shows the two frozen artifacts: a sparse hazard detector and a seven-value steering packet. Validation requests set the detector cutoff and packet strength. Panel B shows the online path. One model pass produces a request summary; if the frozen detector fires, the same packet is added before the model computes its next-token probabilities.}
\label{fig:native-detection-steering-pipeline}
\end{figure}

To build the packet, we give the model one hazardous request followed by a partial refusal ending in \texttt{Assistant: I apologize, but I am}. This provides one example of the internal state we want to reproduce. At the final token, we examine the FFN updates from all 19 layers and select the seven layer-coordinate values with the largest absolute magnitudes. The packet stores each selected layer, coordinate, sign, and value. Dense and MoWE receive packets built separately from their own activations. Once built, the packets do not change.

At test time, every request ends with the neutral partial continuation \texttt{Assistant: I am}. We first apply the sparse linear detector described above. If it classifies the request as benign, we leave the model unchanged. If it classifies the request as hazardous, we add the same seven stored values when the final token passes through their selected layers. The model then computes its next-token probabilities normally. We do not search for new coordinates for each request, and we do not directly force a refusal token.

A single multiplier controls how strongly all seven values are added. For each model, we test a fixed set of multipliers on 20 hazardous validation requests. We choose the smallest multiplier tied for the largest number of refusal predictions, then freeze it before testing.

Two controls check whether the selected signs and coordinate identities matter. First, we reverse the signs of all seven values. Second, we build 100 random packets that keep the same layers, signed values, coordinate count, and strength but place the values at different coordinates. We count a request as steered only when the model's highest-probability next token is one of \{\texttt{unable}, \texttt{sorry}, \texttt{not}, \texttt{afraid}\}, four tokens that can begin common refusal continuations. This is a one-token steering test; it does not evaluate the rest of the response.

\section{Results}
\label{sec:results}

Results proceed in three stages. We first use a controlled task with known latent causes to test whether writing to disjoint coordinates makes those causes easier to locate. We then test whether the same design preserves language-model quality while making the model's own coordinates easier to interpret. Finally, we ask whether those coordinates support direct causal interventions.

\subsection{Known Causes Become Easier to Locate}
\label{sec:results-latent-cause}

Each example contains two noisy bit vectors generated independently from the same unobserved cause. The model sees the first vector and predicts the second, so it must infer the shared cause rather than copy the input.

\begin{table}[H]
\centering
\footnotesize
\setlength{\tabcolsep}{2pt}
\begin{tabular*}{0.99\linewidth}{@{\extracolsep{\fill}}llrrrrr}
\toprule
\shortstack{Model\\(parameters)} & Seed & \shortstack{Binary\\cross-entropy} & \shortstack{Bit\\accuracy} & \shortstack{Exact-vector\\accuracy} & \shortstack{Routing mutual\\information} & \shortstack{Cross-expert\\output overlap} \\
\midrule
Dense (82.8K) & 0 & 0.1059 & 0.9781 & 0.0630 & -- & -- \\
& 1 & 0.1071 & 0.9779 & 0.0569 & -- & -- \\
& Mean & 0.1065 & 0.9780 & 0.0599 & -- & -- \\
\addlinespace[2pt]
MoE (345.5K) & 0 & 0.1051 & 0.9781 & 0.0630 & 2.8621 & 0.01073 \\
& 1 & 0.1063 & 0.9779 & 0.0569 & 3.0803 & 0.00836 \\
& Mean & 0.1057 & 0.9780 & 0.0599 & 2.9712 & 0.00955 \\
\addlinespace[2pt]
MoWE (202.6K) & 0 & 0.1054 & 0.9781 & 0.0630 & 3.0071 & 0.00000 \\
& 1 & 0.1066 & 0.9779 & 0.0569 & 3.0049 & 0.00000 \\
& Mean & 0.1060 & 0.9780 & 0.0599 & \textbf{3.0060} & \textbf{0.00000} \\
\bottomrule
\end{tabular*}
\caption{Prediction and routing on the known-cause task, with unweighted means over the two seeds. Routing mutual information, measured in nats, indicates how much the selected experts reveal about the latent cause. Cross-expert output overlap is the normalized interference between different experts' output contributions; lower is better. These models are matched for task setup and predictive performance, not total parameter count or exact active computation; this experiment is intended to test localization rather than efficiency.}
\label{tab:latent-cause-performance}
\end{table}

As shown in \hyperref[tab:latent-cause-performance]{Table~\ref*{tab:latent-cause-performance}}, all three models have identical bit and exact-vector accuracy, while their binary cross-entropies span only 0.0008. We can therefore compare localization without a meaningful difference in predictive performance.

Both routed models select experts according to the latent cause: their routing carries about three nats of mutual information about it. Because routing specialization is similar, the remaining question is whether the experts preserve that separation when they write to the output representation.

\begin{figure}[H]
\centering
\includegraphics[width=0.86\linewidth]{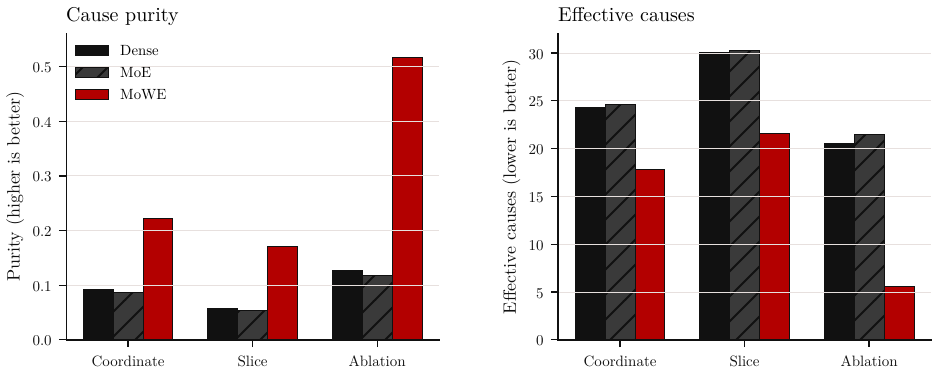}
\caption{Localization of the known causes. Higher purity means that one cause dominates a measurement, while a lower effective-cause count means that it mixes fewer causes. The panels apply these measures to individual coordinates, expert slices, and the prediction changes caused by removing one slice.}
\label{fig:toy-latent-cause-purity}
\end{figure}

\hyperref[fig:toy-latent-cause-purity]{Figure~\ref*{fig:toy-latent-cause-purity}} shows the same pattern at three levels. MoWE has higher purity and fewer effective causes for both individual coordinates and expert slices. In the causal-ablation measurement, removing one MoWE slice affects a narrower set of causes than removing one MoE slice.

The routing mutual information is similar for MoE and MoWE, so both routers learn the cause structure. Their outputs differ: MoE combines expert contributions in shared coordinates, whereas MoWE assigns each expert private coordinates. MoWE consequently has zero cross-expert output overlap by construction, and \hyperref[fig:toy-latent-cause-purity]{Figure~\ref*{fig:toy-latent-cause-purity}} shows that its cause associations are concentrated in fewer of the model's own coordinates. Additional controlled sweeps appear in \hyperref[app:toy-secondary-sweeps]{Appendix~\ref*{app:toy-secondary-sweeps}}.

\subsection{From a Pilot to the Final Language Model}
\label{sec:results-768}
\label{sec:results-scaled-up}

Before training the final models, we needed to choose a stable router recipe and a useful way to scale the architecture. We first screened 25 completed and partial runs at width 512. These experiments support the small auxiliary balance loss and router z-loss used in the later models; the complete screen appears in \hyperref[app:512-scale]{Appendix~\ref*{app:512-scale}}.

A 10-layer model with \(d=768\) provides the first language-model comparison. At matched active compute, dense achieves a best logged cross-entropy 0.0697 nats lower than MoWE.

The pilot uses an earlier readability diagnostic: top-context purity in the native basis minus top-context purity after a matched random rotation. This difference is 0.2570 for MoWE and 0.0329 for dense. A sparse autoencoder with \(16\times\) expansion raises the dense result to 0.1550, still below native MoWE. This pilot diagnostic is distinct from the specificity score used for the final models below; the complete pilot appears in \hyperref[app:lm-768-support]{Appendix~\ref*{app:lm-768-support}}.

We compared four scale-up strategies: increasing residual width, adding transformer layers, increasing each expert's hidden width, and adding depth within the MoWE FFN. Adding transformer layers produced the largest completed improvement and narrowed the dense--MoWE loss gap most. The final comparison therefore keeps \(d=768\) and increases depth to 19 layers. The complete scaling probes appear in \hyperref[app:scaling-probe-details]{Appendix~\ref*{app:scaling-probe-details}}.

\begin{figure}[H]
\centering
\includegraphics[width=0.96\linewidth]{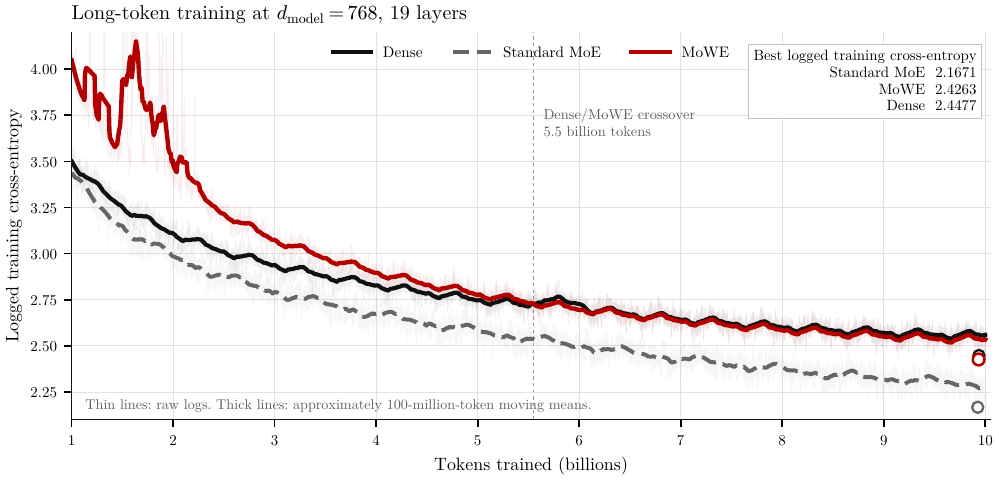}
\caption{Training cross-entropy for one 10-billion-token run of each architecture. Thin lines show logged measurements, thick lines show averages over approximately 100 million tokens, and open circles mark each run's lowest logged value. \hyperref[tab:lm-768-l19-quality]{Table~\ref*{tab:lm-768-l19-quality}} averages two runs per architecture.}
\label{fig:lm-768-l19-10b}
\end{figure}

\begin{table}[H]
\centering
\footnotesize
\setlength{\tabcolsep}{5pt}
\begin{tabular*}{0.78\linewidth}{@{\extracolsep{\fill}}llrr}
\toprule
Model & Seed & \shortstack{Best training\\cross-entropy} & \shortstack{Evaluation\\cross-entropy} \\
\midrule
Dense & 1 & 2.448 & 2.451 \\
& 2 & 2.451 & 2.454 \\
& Mean & 2.449 & 2.452 \\
\addlinespace[2pt]
MoWE & 1 & 2.426 & 2.418 \\
& 2 & 2.423 & 2.417 \\
& Mean & 2.424 & 2.417 \\
\addlinespace[2pt]
Standard MoE & 1 & 2.167 & -- \\
& 2 & 2.230 & -- \\
& Mean & \textbf{2.199} & \textbf{2.220} \\
\bottomrule
\end{tabular*}
\caption{Final 19-layer comparison with individual seed results and unweighted means. The standard-MoE evaluation result was retained only as a two-seed mean, so its seed-wise evaluation values are unavailable. Model sizes and active computation appear in \hyperref[tab:main-lm-configurations]{Table~\ref*{tab:main-lm-configurations}}.}
\label{tab:lm-768-l19-quality}
\end{table}

In the displayed dense--MoWE pair, dense has lower cross-entropy early in training. The smoothed curves in \hyperref[fig:lm-768-l19-10b]{Figure~\ref*{fig:lm-768-l19-10b}} cross at about 5.5 billion tokens, after which MoWE remains lower. This trajectory is consistent with delayed router specialization, but the loss curve alone does not establish that mechanism.

Averaged over two runs, MoWE has lower best training cross-entropy and an evaluation cross-entropy 0.035 nats below dense, as shown in \hyperref[tab:lm-768-l19-quality]{Table~\ref*{tab:lm-768-l19-quality}}. Extended training therefore closes and reverses the gap observed in the 10-layer pilot.

In the fixed-shape NVIDIA RTX A6000 forward-pass benchmark, dense takes 73.27 milliseconds per benchmark iteration, corresponding to 223,600 tokens per second. MoWE takes 74.99 milliseconds, or 218,500 tokens per second. It therefore retains 97.7\% of dense throughput with 2.35\% additional wall-clock time.

Standard MoE achieves lower cross-entropy than both dense and MoWE. Relative to MoWE, it uses 3.3\% more active MACs and 33\% more parameters. We include one standard-MoE checkpoint as a routed baseline in the coordinate-readability evaluation.

\subsubsection{Readable native coordinates across all layers}
\label{sec:results-scaled-up-interpretability}

\begin{figure}[H]
\centering
\includegraphics[width=0.98\linewidth]{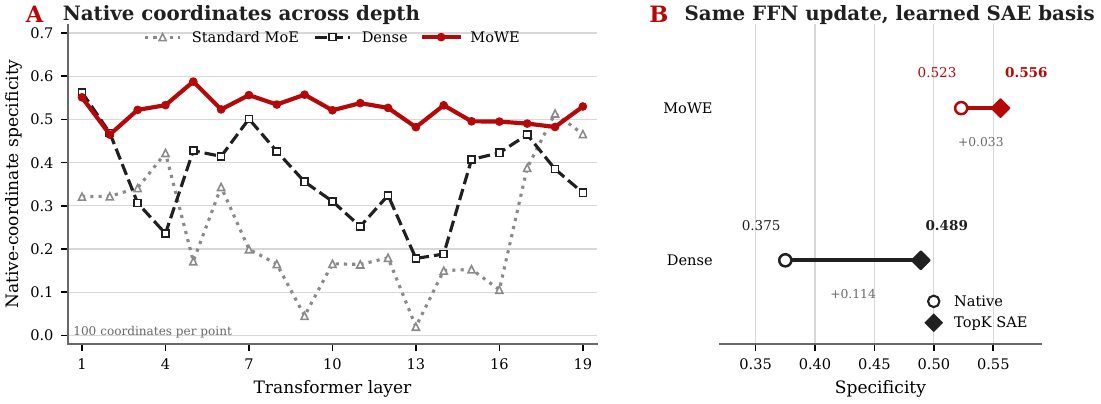}
\caption{Native-coordinate readability across depth and after a site-matched SAE. Panel A shows specificity for 100 native coordinates in each layer. MoWE is higher than dense and standard MoE in 18 of 19 layers. Panel B compares the all-layer means for native coordinates and TopK-SAE features computed from the same FFN update.}
\label{fig:readability-across-layers}
\end{figure}

Each row in \hyperref[tab:lm-all-layer-core]{Table~\ref*{tab:lm-all-layer-core}} contains 100 native coordinates or SAE features from each of 19 layers, for 1,900 in total. The table reports the specificity score defined in \hyperref[sec:method-interpretability]{Section~\ref*{sec:method-interpretability}}. A higher score means that the description is more specific to its source coordinate or feature. The final column shows how much specificity varies across layers; \hyperref[fig:readability-across-layers]{Figure~\ref*{fig:readability-across-layers}} shows the corresponding layer-wise pattern and site-matched SAE comparison.

\begin{table}[H]
\centering
\footnotesize
\setlength{\tabcolsep}{4pt}
\begin{tabular*}{0.96\linewidth}{@{\extracolsep{\fill}}llrrrr}
\toprule
Model & Representation & Raw purity & Shuffled floor & Specificity & \shortstack{Across-layer\\standard deviation} \\
\midrule
\multicolumn{6}{l}{\textit{Native-coordinate audit}} \\
Dense & Native FFN update & 0.573 & 0.317 & 0.375 & 0.102 \\
Dense & \shortstack{Random rotation of\\native FFN update} & 0.567 & 0.308 & 0.375 & 0.088 \\
Standard MoE & Native FFN update & 0.630 & 0.495 & 0.268 & 0.138 \\
Standard MoE & \shortstack{Random rotation of\\native FFN update} & 0.633 & 0.484 & 0.289 & 0.126 \\
MoWE & Native FFN update & 0.601 & 0.162 & \textbf{0.523} & \textbf{0.030} \\
MoWE & \shortstack{Random rotation of\\native FFN update} & 0.633 & 0.403 & 0.384 & 0.130 \\
\addlinespace[3pt]
\multicolumn{6}{l}{\textit{TopK-SAE audit at the same FFN-update site}} \\
Dense & TopK SAE & 0.570 & 0.159 & 0.489 & 0.065 \\
Dense & Random-rotation control & 0.572 & 0.298 & 0.390 & 0.101 \\
MoWE & TopK SAE & 0.619 & 0.143 & \textbf{0.556} & 0.046 \\
MoWE & Random-rotation control & 0.629 & 0.393 & 0.388 & 0.110 \\
\bottomrule
\end{tabular*}
\caption{Top-context purity across 19 layers, using 100 live coordinates or SAE features per layer. Specificity corrects raw purity for the shuffled-description floor; higher is better. Both panels evaluate the same FFN update immediately before residual addition. The native audit reads its original coordinates; the SAE audit learns a TopK dictionary from that same vector.}
\label{tab:lm-all-layer-core}
\end{table}

Three findings stand out. First, MoWE has the highest native-coordinate specificity: 0.523, compared with 0.375 for dense and 0.268 for standard MoE. Relative to standard MoE, the difference is \(+0.255\), with a 95\% confidence interval (CI) of \([+0.185,+0.333]\), and MoWE is higher in 18 of 19 layers. MoWE is also more consistent across depth: its across-layer standard deviation is 0.030, versus 0.102 for dense and 0.138 for standard MoE.

Second, the site-matched SAE adds more to dense than to MoWE. Dense rises from 0.375 in its native coordinates to 0.489 with a TopK SAE, while MoWE rises from 0.523 to 0.556. Native MoWE is therefore comparable to the dense SAE: the 95\% CI for their difference is \([-0.002,+0.070]\). The MoWE SAE also exceeds the dense SAE by \(+0.067\), with a 95\% CI of \([+0.027,+0.110]\). Much of the structure that the SAE adds to dense is already visible in MoWE's original coordinates.

Third, the rotation controls separate readable axes from broadly describable activity. Standard MoE has raw purity of 0.630, but its high shuffled floor of 0.495 leaves specificity of only 0.268. Its random rotation scores a similar 0.289. MoWE instead scores 0.523 in its native basis and 0.384 after rotation. Sparse routing alone therefore does not produce readable output coordinates; the improvement is associated with MoWE's private, disjoint outputs. Complete controls and confidence intervals appear in \hyperref[app:lm-all-layer-controls]{Appendix~\ref*{app:lm-all-layer-controls}}.

\subsection{Readable Coordinates Can Be Used Directly}
\label{sec:results-case-study}

The aggregate evaluation shows that MoWE coordinates receive more distinctive descriptions. We now test whether those coordinates support causal analysis and intervention. The first example traces a factual prediction; the second stores and reuses native coordinates during inference.

\subsubsection{Following and reusing a measurement path}

The final MoWE model predicts \texttt{degrees} after \emph{A right angle is 90} with probability 0.715. A precomputed coordinate atlas labels four active coordinates with related descriptions: numeric literals, measurement numbers, quantity and measurement, and measurements and distances. These are independent summaries of each coordinate's strongest contexts, not names for formal stages of arithmetic. In the notation \(L3.u402\), \(L3\) denotes layer 3 and \(u402\) denotes coordinate 402. The resulting candidate path is:
\[
L3.u402 \rightarrow L7.u424 \rightarrow L8.u757 \rightarrow L13.u274 \rightarrow \texttt{degrees}.
\]

\begin{figure}[H]
\centering
\includegraphics[width=0.98\linewidth]{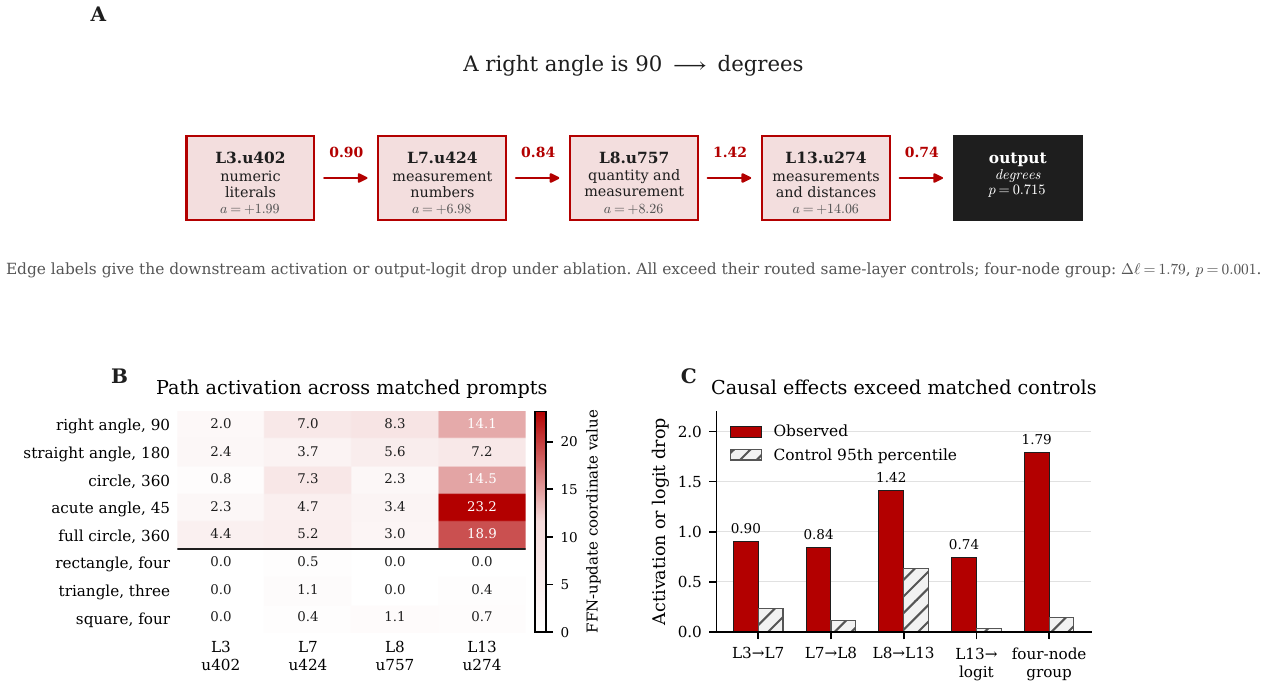}
\caption{A native-coordinate path for an angle completion. Panel A shows the selected coordinates and the downstream activation change caused by setting each one to zero. Panel B compares their activations on angle and side-count prompts. Panel C compares each selected causal effect with same-layer controls; the final comparison ablates all four coordinates against 1,000 layer-matched coordinate groups.}
\label{fig:mowe-19l-angle-path}
\end{figure}

The four coordinates pass two checks in \hyperref[fig:mowe-19l-angle-path]{Figure~\ref*{fig:mowe-19l-angle-path}}. First, their activations match the labels: they are active on right-angle, straight-angle, circle, and acute-angle prompts, but small or inactive on rectangle, triangle, and square side-count prompts. Second, setting each coordinate to zero changes its downstream target more than at least 95\% of the active alternatives from the same layer.

Setting all four coordinates to zero lowers the \texttt{degrees} logit by 1.790. Among 1,000 layer-matched coordinate groups, the 95th-percentile drop is 0.148 and the empirical \(p\)-value is 0.001. The prediction remains \texttt{degrees}, but its probability falls from 0.715 to 0.670, showing that the path is influential without being the model's only route to the answer.

We next ask whether angle-related coordinates can influence another prompt. This is a test of semantic control, not an attempt to improve the answer: \texttt{sides} is the correct completion of \emph{A rectangle has four}. Before intervention, the model predicts \texttt{sides} with probability 0.2257, while \texttt{degrees} ranks 166th. We select the seven largest positive coordinate values from the angle prompt's FFN updates and add scaled versions of those values at the final position of the rectangle prompt.

\begin{figure}[H]
\centering
\includegraphics[width=0.92\linewidth]{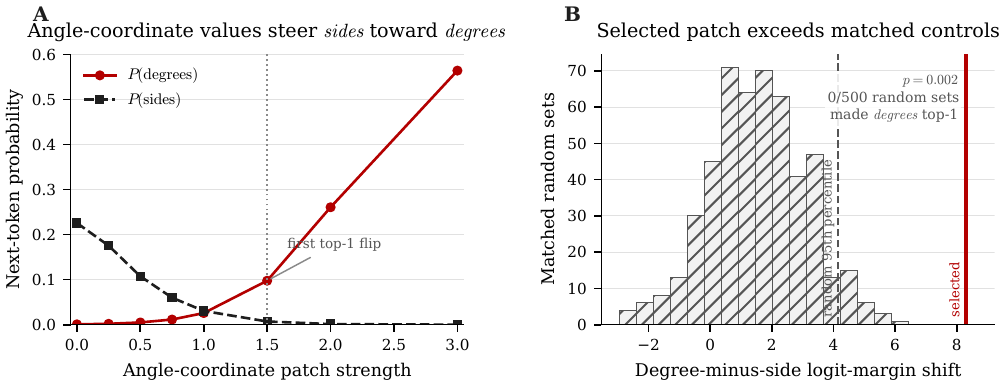}
\caption{Reusing angle-related coordinate values. Panel A shows how next-token probabilities change as the multiplier on the stored seven-coordinate intervention increases. Panel B compares the selected intervention with 500 controls matched for layer, coordinate count, sign, and amplitude.}
\label{fig:mowe-19l-angle-patch}
\end{figure}

At multiplier 1.5, \texttt{degrees} becomes the top prediction: its probability rises to 0.0976 while \texttt{sides} falls to 0.0071. None of 500 matched random coordinate sets produces the same top-1 change (\(p=0.002\)). This selected, post-hoc example demonstrates that a labeled native-coordinate set can be stored and reused without first translating it through a learned dictionary.

Because the prompt family, coordinate set, and displayed intervention strength were selected during case-study construction, these empirical \(p\)-values compare the chosen intervention with matched random controls but are not corrected for the preceding selection process.

\subsubsection{Detecting and steering at inference time}

The second case study evaluates a two-stage policy. A detector first classifies each request as hazardous or benign. If it detects a hazard, the model receives a fixed seven-coordinate intervention. We evaluate the detector and intervention separately before combining them. The test set contains 100 hazardous requests and 100 benign controls built from five matched templates that are unseen during fitting and threshold selection. This is a matched-template proof of concept, not a general safety classifier. We report accuracy and area under the receiver operating characteristic curve (AUC).

The single-prompt top-7 detector selects seven coordinates from one reference bomb request. Contrastive top-7 selects seven coordinates that distinguish hazardous fitting requests from benign controls. The sparse linear gate fits a classifier over all native-coordinate activations.

\begin{table}[H]
\centering
\footnotesize
\setlength{\tabcolsep}{3pt}
\begin{tabularx}{0.98\linewidth}{@{}l>{\raggedright\arraybackslash}Xccccc@{}}
\toprule
Model & Detector & \shortstack{Detector\\coordinates} & \shortstack{Hazardous requests\\detected} & False alarms & Accuracy & AUC \\
\midrule
MoWE & Single-prompt top 7 & 7 & 41/100 & 11/100 & 65.0\% & 0.722 \\
Dense & Single-prompt top 7 & 7 & 3/100 & 11/100 & 46.0\% & 0.444 \\
MoWE & Contrastive top-7 & 7 & 84/100 & 24/100 & 80.0\% & 0.897 \\
Dense & Contrastive top-7 & 7 & 85/100 & 48/100 & 68.5\% & 0.844 \\
MoWE & Sparse linear gate & \textbf{15} & \textbf{99/100} & 8/100 & \textbf{95.5\%} & \textbf{0.999} \\
Dense & Sparse linear gate & 28 & 86/100 & \textbf{1/100} & 92.5\% & 0.987 \\
\bottomrule
\end{tabularx}
\caption{Request detection from native coordinates. Counts are out of 100 hazardous and 100 matched benign requests. False alarms are benign requests classified as hazardous. Accuracy is measured over all 200 requests. An AUC of 1 is perfect and 0.5 is chance.}
\label{tab:native-coordinate-detection}
\end{table}

The sparse linear gates perform best overall. MoWE uses fewer coordinates and has higher accuracy and AUC, while dense produces fewer false alarms.

To test the intervention itself, we select seven coordinates from one example formatted to continue with a refusal and store their signed values as a fixed packet. Selection occurs once, before evaluation. The same packet is then applied to every hazardous test request, with no prompt-specific coordinate search and no token forced directly at the output layer.

Packet strength is selected separately for each model on 20 validation requests. From a fixed grid, we choose the smallest multiplier that produces the maximum number of refusal-family predictions. This procedure selects strength 1.50 for MoWE and 0.10 for dense.

\begin{table}[H]
\centering
\footnotesize
\setlength{\tabcolsep}{2.5pt}
\begin{tabular*}{0.99\linewidth}{@{\extracolsep{\fill}}lcccccc@{}}
\toprule
Model & Strength & \shortstack{Refusals\\before} & \shortstack{Refusals after\\packet} & \shortstack{Refusals after\\sign reversal} & \shortstack{Matched-random refusals\\mean \(\pm\) standard deviation} & Empirical \(p\) \\
\midrule
MoWE & 1.50 & 0/100 & \textbf{100/100} & 0/100 & \(6.8\pm15.8\) & 0.0099 \\
Dense & 0.10 & 2/100 & 2/100 & 2/100 & \(3.2\pm2.9\) & 0.7327 \\
\bottomrule
\end{tabular*}
\caption{Number of hazardous requests, out of 100, whose next-token prediction belongs to the refusal family. Strength is the multiplier applied to the stored packet. Random packets preserve layer, signed value, and coordinate count while changing coordinate identity.}
\label{tab:native-coordinate-steering}
\end{table}

The fixed MoWE packet changes all 100 next-token predictions to a refusal-family token. Reversing its signs removes the effect. Matched random packets produce \(6.8\pm15.8\) refusals on average, giving empirical \(p=0.0099\).

The dense seven-coordinate packet produces no new refusals and is indistinguishable from its controls: two refusals are observed, compared with \(3.2\pm2.9\) for random packets (\(p=0.7327\)). Dense coordinates therefore support a strong detector, but this compact dense packet does not steer the model. MoWE supports both detection and steering. Six of its seven packet coordinates have atlas labels involving auxiliaries, copulas, contractions, negation, or nearby function-word behavior, consistent with the continuation \texttt{I am not}.

Finally, we combine each model's pre-fitted detector with its stored packet. At inference time, the detector scores the request using native-coordinate activations. The packet is applied only when that score exceeds the fixed threshold.

\begin{table}[H]
\centering
\footnotesize
\setlength{\tabcolsep}{5pt}
\begin{tabular*}{0.82\linewidth}{@{\extracolsep{\fill}}lrr}
\toprule
Measure & MoWE & Dense \\
\midrule
Detector coordinates & \textbf{15} & 28 \\
Packet coordinates & 7 & 7 \\
Hazardous requests detected & \textbf{99/100} & 86/100 \\
Detected hazards converted & \textbf{99/99} & 0/86 \\
Final hazardous refusals & \textbf{99/100} & 2/100 \\
Benign requests detected & 8/100 & \textbf{1/100} \\
New benign refusals & 8/100 & \textbf{0/100} \\
Final benign refusals & 9/100 & \textbf{0/100} \\
Accuracy & \textbf{95.0\%} & 51.0\% \\
\bottomrule
\end{tabular*}
\caption{The pre-fitted detector and stored packet on 100 hazardous and 100 matched benign requests. A correct outcome is a refusal-family next token for a hazardous request and a non-refusal token for a benign request. Accuracy is measured over all 200 requests.}
\label{tab:native-coordinate-end-to-end}
\end{table}

\begin{figure}[H]
\centering
\includegraphics[width=0.98\linewidth]{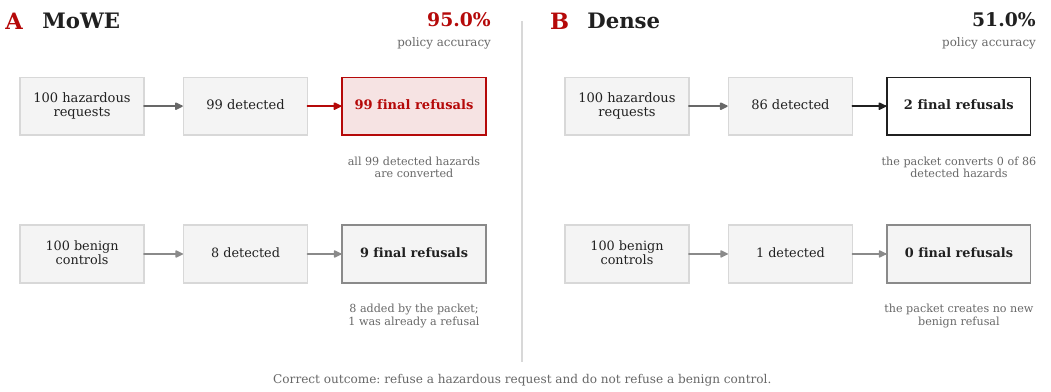}
\caption{End-to-end outcomes for the frozen detector and stored packet. Each panel follows the 100 hazardous requests and 100 matched benign controls through the corresponding policy. A correct outcome is a refusal-family next token for a hazardous request and a non-refusal next token for a benign control.}
\label{fig:end-to-end-safety-outcomes}
\end{figure}

\hyperref[tab:native-coordinate-end-to-end]{Table~\ref*{tab:native-coordinate-end-to-end}} gives the exact counts, while \hyperref[fig:end-to-end-safety-outcomes]{Figure~\ref*{fig:end-to-end-safety-outcomes}} shows how requests move through the two policies. End to end, MoWE produces refusals for 99 of 100 hazardous requests. One benign control already produces a refusal-family token before intervention, and the detector and packet add eight more. The final nine benign refusals yield 190 correct outcomes out of 200 (95.0\%). Dense detects 86 hazardous requests, but its compact packet converts none of them; including two baseline refusals, dense produces 102 correct outcomes out of 200 (51.0\%).

Together, these experiments connect coordinate readability to model behavior. Coordinate labels identify candidate features, ablations test their causal importance, and stored coordinate sets can be reused without prompt-specific search. The scope remains narrow: all 99 successful interventions change only the next token to \texttt{not}, and we do not evaluate the remainder of the generated response.

\section{Discussion}
\label{sec:discussion}

MoWE gives every routed expert a stable output address. Training decides what each address represents, but different experts cannot contribute to the same coordinates within one FFN update. The experiments show that this simple constraint produces more localized structure in the model's native coordinates.

\paragraph{What writing to disjoint coordinates changes.}
The known-cause task separates the effect of routing from the effect of writing to disjoint coordinates. Standard MoE and MoWE both route examples according to the hidden cause, but only MoWE gives each expert a separate output coordinate. The language-model comparison shows the same distinction in a less controlled setting. Standard MoE reaches the lowest training and evaluation loss, but its native output coordinates are no more specific than a random rotation. MoWE combines conditional routing with stable output addresses and develops substantially more specific native coordinates.

\paragraph{Relationship to sparse autoencoders.}
Sparse autoencoders remain useful, but MoWE changes how much structure is visible before post-hoc decomposition. At the same FFN-update site, dense improves from 0.375 in its native coordinates to 0.489 with a TopK SAE. MoWE starts higher at 0.523 and reaches 0.556 with its own SAE. The larger gain for dense shows what its learned dictionary adds. The smaller gain for MoWE shows that much of this structure is already present in the model's own coordinates. An SAE can still give MoWE a larger or finer feature dictionary when needed.

\paragraph{Prediction quality and systems cost.}
MoWE stores more expert weights than dense but executes only the selected experts for each token. Its loss is higher early in training, but both 10-billion-token MoWE runs finish with lower evaluation cross-entropy than their active-compute-matched dense runs. Standard MoE obtains the lowest mean training and evaluation loss while using 3.3\% more active MACs than MoWE. In a fixed-shape forward benchmark, the optimized MoWE implementation retains 97.7\% of dense throughput. The architectures therefore have different strengths: standard MoE is the strongest predictor tested, while MoWE is the routed model that develops readable native output coordinates.

\paragraph{Native coordinates as an intervention interface.}
The case studies show why native coordinates can be useful. In the geometry example, coordinate labels reveal a path that supports \emph{degrees} after an angle prompt. Reusing angle-related coordinate values then deliberately shifts a separate rectangle prompt toward \emph{degrees}. In the request experiment, a sparse linear gate detects when to intervene and a stored seven-coordinate packet supplies the intervention. Dense coordinates support a strong detector, but the matched dense packet changes no new predictions. In this setting, MoWE coordinates act as readable and reusable control handles.

\subsection{Limitations and Next Steps}
\label{sec:limitations}

The loss comparison contains two runs per architecture, while the all-layer interpretability study evaluates one checkpoint from each architecture. The SAE audit includes only live features. The intervention study measures one next-token decision in one request family.

The next tests are larger models, additional data mixtures and seeds, full-generation steering, and broader human evaluation. \hyperref[app:extended-limitations]{Appendix~\ref*{app:extended-limitations}} describes these extensions in detail.

\section{Conclusion}
\label{sec:conclusion}

Mixture of Wide Experts changes where routed experts contribute to the FFN update. Every selected expert reads the full representation but returns values only for a fixed part of that update. On a controlled task, this rule localizes known causes that standard expert routing leaves distributed. In the final language models, standard MoE achieves the lowest mean training and evaluation loss, but its native coordinates are no more readable than a random rotation. MoWE achieves lower evaluation cross-entropy than dense in both 10-billion-token runs and produces substantially more specific native coordinates than either baseline in the automated evaluation.

The native-coordinate representation supports causal tracing, transfer across related prompts, request-category detection, and fixed next-token steering packets. These results isolate the contribution of writing to disjoint coordinates. Sparse routing supplies conditional computation; stable output addresses allow useful feature structure to appear directly in the coordinates used by the model itself.

\clearpage
\begingroup
\small
\setlength{\bibsep}{3pt}
\bibliographystyle{plainnat}
\bibliography{references}
\endgroup

\clearpage
\appendix
\section*{Appendix Guide}

The appendices provide a second path through the paper for readers who want to follow one question in depth. \hyperref[app:toy-secondary-sweeps]{Appendix~\ref*{app:toy-secondary-sweeps}} tests causal localization in additional controlled settings, while \hyperref[app:512-scale]{Appendix~\ref*{app:512-scale}} reconstructs the router-stabilization search that produced the final recipe. \hyperref[app:method-metric-details]{Appendix~\ref*{app:method-metric-details}} gives detailed experimental protocols and metric definitions. \hyperref[app:lm-support]{Appendix~\ref*{app:lm-support}} shows where the language-model interpretability signal first appears and develops the full SAE and basis controls. \hyperref[app:scaling-probe-details]{Appendix~\ref*{app:scaling-probe-details}} compares every completed scaling direction, and \hyperref[app:extended-limitations]{Appendix~\ref*{app:extended-limitations}} describes the main open questions and the experiments that would answer them.

\section{Additional Evidence for Causal Localization}
\label{app:toy-secondary-sweeps}

The main controlled experiment asks whether giving each expert a private set of residual coordinates localizes a known latent cause. The two experiments in this appendix test the same idea under two changes. The first replaces binary prediction with clustered sparse reconstruction. The second varies the strength of router regularization. Together, they test whether localization survives a different target format and whether strong balancing suppresses useful specialization.

\subsection{A Second Controlled Setting: Clustered Sparse Reconstruction}
\label{app:toy-clustered-reconstruction}

Each input contains 128 binary features, divided into eight latent clusters of 16 features. For every example, one cluster is selected as active. Features in that cluster turn on independently with probability \(p_\mathrm{within}\in\{0.20,0.40,0.60,0.80,1.00\}\); every background feature turns on with probability 0.01. The target is the complete binary input vector.

Every model has a dense encoder, three residual blocks, a dense decoder, and residual width 128. Routed models use \(G=8\) experts, select \(k=2\) experts, and use expert hidden width \(H=16\) and MoWE slice width \(s=16\). The dense hidden width is 32, equal to the total active routed width \(kH\). We train with AdamW for 2,000 steps using learning rate \(10^{-3}\), weight decay \(10^{-4}\), gradient clipping at 1.0, batch size 1,024, and evaluation batch size 8,192.

MoWE therefore uses 92,200 parameters, compared with 137,900 for MoE, a reduction of 33.1\%.

\begin{table}[H]
\centering
\footnotesize
\setlength{\tabcolsep}{3pt}
\begin{tabular*}{0.96\linewidth}{@{\extracolsep{\fill}}lrrrrr}
\toprule
\(p_\mathrm{within}\) & \shortstack{Dense exact\\match} & \shortstack{MoE exact\\match} & \shortstack{MoWE exact\\match} & \shortstack{MoE cross-expert\\interference} & \shortstack{MoWE cross-expert\\interference} \\
\midrule
0.20 & 0.8995 & 0.9370 & \textbf{0.9585} & 0.04698 & \textbf{0.00000} \\
0.40 & 0.8699 & 0.9208 & \textbf{0.9388} & 0.04250 & \textbf{0.00000} \\
0.60 & 0.8553 & 0.9163 & \textbf{0.9359} & 0.03518 & \textbf{0.00000} \\
0.80 & 0.8469 & \textbf{0.9127} & 0.9046 & 0.03076 & \textbf{0.00000} \\
1.00 & \textbf{0.8397} & 0.7751 & 0.8134 & 0.02016 & \textbf{0.00000} \\
\bottomrule
\end{tabular*}
\caption{Clustered sparse reconstruction. MoWE uses 33.1\% fewer parameters than MoE and has higher exact-match accuracy at four of the five sweep points. Dense is best when \(p_\mathrm{within}=1.00\). MoWE cross-expert output interference is zero by construction throughout the sweep.}
\label{tab:clustered-sparse-reconstruction}
\end{table}

\hyperref[tab:clustered-sparse-reconstruction]{Table~\ref*{tab:clustered-sparse-reconstruction}} shows a consistent tradeoff rather than universal dominance. MoWE has higher exact-match accuracy than MoE for \(p_\mathrm{within}=0.20,0.40,0.60,\) and \(1.00\). MoE is slightly better at \(p_\mathrm{within}=0.80\), and dense is best at \(p_\mathrm{within}=1.00\). Binary cross-entropy gives the same MoWE-versus-MoE ordering at all five points. Thus, despite using fewer parameters, MoWE remains competitive on reconstruction while ensuring that different experts do not write to overlapping residual coordinates.

\subsection{Why Light Router Stabilization Is Enough}
\label{app:toy-router-stabilization}

This screen uses 256 binary features arranged into 16 clusters of 16, with one active cluster per example. Within-cluster features activate with probability 0.85, while background features activate with probability 0.01. The MoWE model has residual width 256, four residual blocks, \(G=16\), \(k=2\), slice width 16, expert hidden width 8, and 288,800 parameters. Only the auxiliary load-balancing coefficient and router z-loss coefficient vary; the optimizer and all other training settings remain fixed. The experiment asks how much stabilization is needed to recover routing aligned with the latent clusters without sacrificing reconstruction.

The sweep shows the same pattern as the language-model ablations: a very small balance coefficient improves alignment between routing and latent clusters, whereas stronger balancing spreads traffic across more experts and lowers reconstruction accuracy.

\begin{table}[H]
\centering
\footnotesize
\setlength{\tabcolsep}{3pt}
\begin{tabular*}{0.96\linewidth}{@{\extracolsep{\fill}}lrrrrr}
\toprule
Variant & \shortstack{Exact-match\\accuracy} & \shortstack{Binary cross-\\entropy} & \shortstack{Routing mutual\\information} & \shortstack{Effective experts\\per cluster} & \shortstack{Cross-expert\\interference} \\
\midrule
No auxiliary loss & 0.9725 & \textbf{0.001151} & 0.0088 & 1.90 & 0.00000 \\
Auxiliary coefficient \(10^{-4}\) & \textbf{0.9744} & 0.001183 & \textbf{1.5645} & 4.13 & 0.00000 \\
Auxiliary \(10^{-4}\) + z-loss \(10^{-3}\) & 0.9714 & 0.001246 & 1.3327 & 5.27 & 0.00000 \\
Auxiliary coefficient \(10^{-2}\) & 0.9669 & 0.001257 & 0.6335 & 9.51 & 0.00000 \\
Auxiliary coefficient \(5\times10^{-2}\) & 0.9467 & 0.001929 & 0.3097 & 12.32 & 0.00000 \\
\bottomrule
\end{tabular*}
\caption{MoWE router-stabilization sweep. Routing mutual information measures how strongly selected experts identify the active latent cluster. An auxiliary coefficient of \(10^{-4}\) gives the highest exact-match accuracy and much stronger routing alignment. Coefficients of \(10^{-2}\) and \(5\times10^{-2}\) spread routing across more experts but degrade reconstruction.}
\label{tab:toy-router-stabilization}
\end{table}

\hyperref[tab:toy-router-stabilization]{Table~\ref*{tab:toy-router-stabilization}} separates reconstruction quality from routing structure. Without an auxiliary loss, MoWE reconstructs well but has routing mutual information of only 0.0088, so accurate reconstruction alone does not imply routing aligned with the latent clusters. An auxiliary coefficient of \(10^{-4}\) gives both the best exact-match accuracy and much higher routing mutual information. Coefficients of \(10^{-2}\) and \(5\times10^{-2}\) activate more experts per cluster but reduce both routing alignment and task performance. Light stabilization is therefore sufficient; the router need not approach uniform expert use.

\section{From Unstable Routing to the Final Recipe}
\label{app:512-scale}

This appendix explains how we selected the light router-stabilization recipe used in the main experiments. The study includes nineteen completed one-billion-token stabilization and baseline runs, followed by six matched-batch 300-million-token router-curriculum screens and controls; one screen stopped early. A small auxiliary balance term combined with router z-loss stabilizes training. Forcing broad expert use does not improve cross-entropy and often makes it worse. All reported runs use corrected, fixed rotary position embeddings (RoPE) \citep{su2021roformer}; runs completed before this correction are excluded. Because these experiments use a science-only training stream, their cross-entropy values should not be compared directly with the later science-and-code experiments.

\subsection{Light stabilization is enough}
\label{app:512-light-stabilization}

The router-stability ablation uses residual width \(d=512\), 10 transformer layers, eight attention heads, sequence length 1,024, and the science-only stream. MoWE has \(G=512\) experts, selects \(k=26\) per token, and gives each expert hidden width \(H=8\). The dense baseline has feed-forward width \(m=310\), chosen to match MoWE's active multiply--accumulate operations per token. Router health is reported at layer 9. The effective-expert count is the entropy-based number of experts in use; maximum load is the busiest expert's load divided by its load under uniform routing. Some ablations also use a load-dependent routing bias that changes expert selection without changing the gate weights applied to selected experts.

\begin{figure}[H]
\centering
\includegraphics[width=0.86\linewidth]{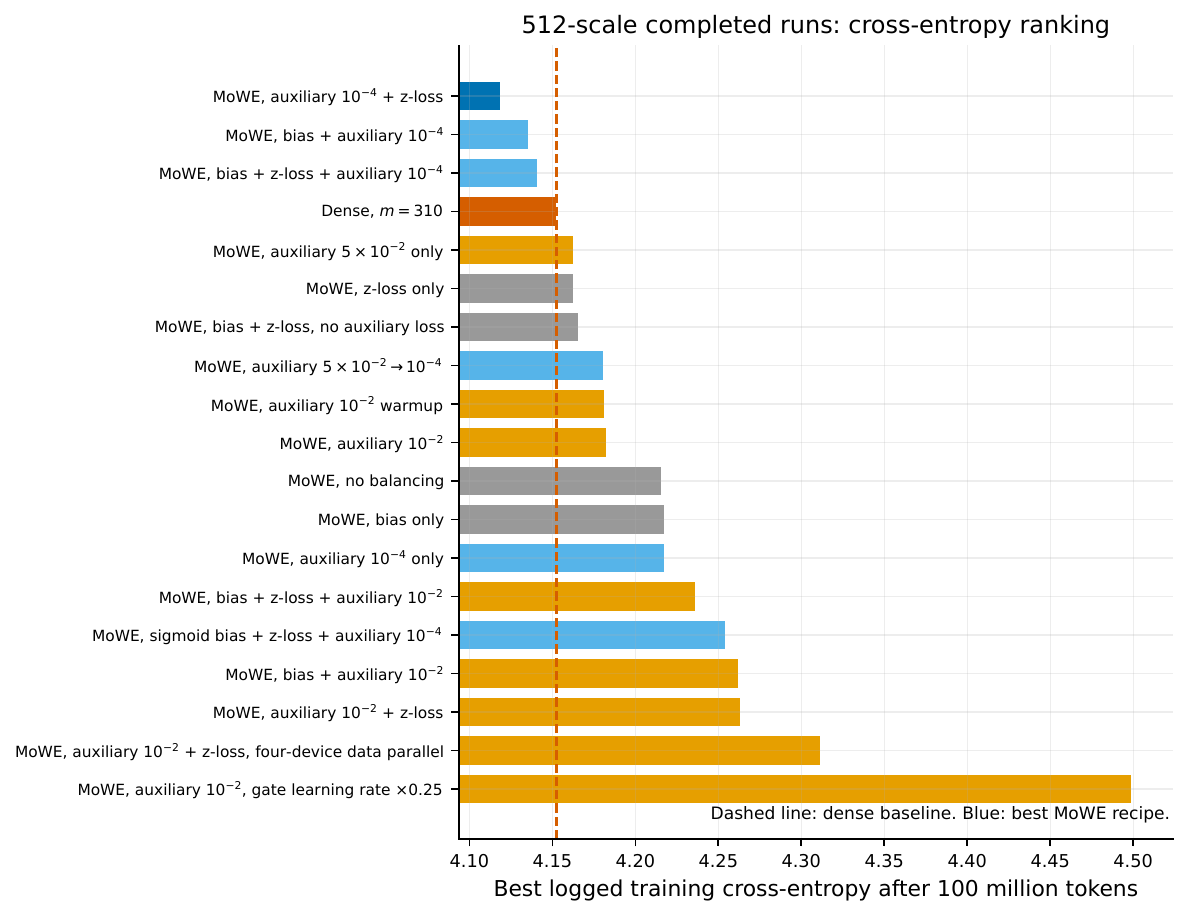}
\caption{All completed 512-scale one-billion-token runs ranked by best logged training cross-entropy, measured after the first 100 million tokens.}
\label{fig:lm-512-ranking}
\end{figure}

\begin{table}[H]
\centering
\footnotesize
\setlength{\tabcolsep}{3pt}
\begin{tabular*}{0.96\linewidth}{@{\extracolsep{\fill}}lrrr}
\toprule
Run & \shortstack{Best training\\cross-entropy} & \shortstack{Layer-9 effective\\expert count} & \shortstack{Layer-9 maximum load\\relative to uniform} \\
\midrule
MoWE auxiliary \(10^{-4}\) + z-loss & \textbf{4.1186} & 474.3 & 2.73 \\
MoWE bias + auxiliary \(10^{-4}\) & 4.1352 & 472.0 & 2.97 \\
MoWE bias + z-loss + auxiliary \(10^{-4}\) & 4.1405 & 469.3 & 2.96 \\
Dense \(m=310\) & 4.1524 & -- & -- \\
MoWE auxiliary \(5\times10^{-2}\) only & 4.1624 & 500.7 & 1.80 \\
MoWE z-loss only & 4.1626 & 28.0 & 19.69 \\
\bottomrule
\end{tabular*}
\caption{Top 512-scale corrected fixed-RoPE one-billion-token runs by best logged training cross-entropy after 100 million tokens.}
\label{tab:lm-512-top-runs}
\end{table}

\hyperref[fig:lm-512-ranking]{Figure~\ref*{fig:lm-512-ranking}} and \hyperref[tab:lm-512-top-runs]{Table~\ref*{tab:lm-512-top-runs}} show that the best 512-scale MoWE run combines an auxiliary load-balancing coefficient of \(10^{-4}\) with router z-loss and reaches slightly lower cross-entropy than the active-compute-matched dense baseline. The ablation also separates load balance from language-model quality. Strong auxiliary loss spreads traffic broadly but raises cross-entropy. Router z-loss alone concentrates routing on 28 effective experts, yet remains close to the dense baseline in cross-entropy. Bias-only variants rank lower. We therefore use light stabilization that allows specialization without requiring uniform expert use.

\subsection{Forcing broad routing does not solve optimization}
\label{app:512-router-curriculum}

The shorter experiments are matched-batch screens evaluated by their best logged cross-entropy over at most 300 million tokens. They use the same 512-scale architecture, 32,768 tokens per optimizer update, and a learning-rate schedule configured for a one-billion-token run. We stop them at 300 million tokens to save compute.

During random-routing warmup, the gate remains untrained for 150 million tokens. Its initially unspecialized selections therefore remain approximately uniform before gate training begins. High-\(k\) warmup reduces the number of selected experts from 104 to the base value of 26 over 200 million tokens. The gentle entropy-floor curriculum uses coefficient \(10^{-5}\) and raises its target effective-expert count from 64 to 400 over 250 million tokens, with a \(2\times\) multiplier from layer 7 onward. The combined curriculum applies all three interventions and raises its entropy target from 128 to 400 over 250 million tokens. Its loss began to diverge, so we stopped it at 191.7 million tokens.

\begin{table}[H]
\centering
\footnotesize
\setlength{\tabcolsep}{2pt}
\begin{tabularx}{0.99\linewidth}{@{}>{\raggedright\arraybackslash}Xcccc@{}}
\toprule
Run & \shortstack{Tokens\\(billions)} & \shortstack{Best training\\cross-entropy} & \shortstack{Layer-9 effective\\expert count} & \shortstack{Layer-9 maximum load\\relative to uniform} \\
\midrule
MoWE auxiliary \(10^{-4}\) screen control & 0.2998 & \textbf{4.7128} & 403.7 & 5.25 \\
Entropy-floor gentle & 0.2998 & 4.7269 & 401.6 & 5.25 \\
High-\(k\) warmup & 0.2998 & 4.9930 & 396.3 & 6.25 \\
Random warmup & 0.2998 & 5.7652 & 297.3 & 11.27 \\
Combined entropy-floor, high-\(k\), and random warmup (partial) & 0.1917 & 5.8158 & 227.1 & 12.33 \\
Dense screen control & 0.2998 & 5.5867 & -- & -- \\
\bottomrule
\end{tabularx}
\caption{Matched-batch router-curriculum screen. Best cross-entropy is measured over each run's available trajectory. The combined curriculum stopped at 191.7 million tokens after its loss began to diverge; the other runs reached approximately 300 million tokens.}
\label{tab:lm-512-300m-screen}
\end{table}

\hyperref[tab:lm-512-300m-screen]{Table~\ref*{tab:lm-512-300m-screen}} shows that broad expert use is not sufficient for good optimization. The gentle entropy floor maintains a high effective-expert count but does not improve on the auxiliary-\(10^{-4}\) control. High-\(k\) warmup, random warmup, and their combination all produce higher cross-entropy. The dense row is a screening reference, not the dense baseline used in the completed one-billion-token comparison.

\subsection{Standard MoE improves loss, not residual-coordinate readability}
\label{app:512-moe-diagnostic}

The local standard-MoE diagnostic uses \(d=512\), 10 layers, eight attention heads, \(G=512\), \(k=26\), \(H=8\), and the science-only stream. Under the parameter-counting convention used by the experiment scripts, standard MoE has approximately 101.8 million parameters and 52.5 million active MACs per token. MoWE has 80.8 million parameters and 51.5 million active MACs per token. The dense baseline has 41.0 million parameters and is matched to MoWE in active compute.

\begin{table}[H]
\centering
\footnotesize
\setlength{\tabcolsep}{4pt}
\begin{tabular*}{0.96\linewidth}{@{\extracolsep{\fill}}lrrl}
\toprule
Model & \shortstack{Parameters\\(millions)} & \shortstack{Best training\\cross-entropy} & Role \\
\midrule
MoE 512 & 101.8 & \textbf{4.0331} & Standard routed baseline \\
MoWE 512 auxiliary \(10^{-4}\) + z-loss & 80.8 & 4.1186 & Stabilized MoWE model \\
Dense 512 \(m=310\) & 41.0 & 4.1524 & Active-compute baseline \\
\bottomrule
\end{tabular*}
\caption{Training loss on the 512-scale science corpus. Standard MoE has more parameters than MoWE and reaches the lowest cross-entropy.}
\label{tab:lm-512-loss}
\end{table}

Standard MoE provides the expected loss advantage at this scale \citep{fedus2022switch,muennighoff2024olmoe}. We next ask whether this advantage also produces readable native output coordinates. At layer 6, we sample 100 coordinates from each model's feed-forward network update, the vector added back to the residual stream, and evaluate top-context purity. The judge describes each coordinate from its highest-activating contexts, then scores that description on separate high-activation contexts.

\begin{table}[H]
\centering
\footnotesize
\setlength{\tabcolsep}{3pt}
\begin{tabular*}{0.96\linewidth}{@{\extracolsep{\fill}}lrrrrrr}
\toprule
Model & Raw purity & \shortstack{95\% confidence\\interval} & Median & \shortstack{Fraction\\\(>0.7\)} & \shortstack{Shuffled\\floor} & \shortstack{Coordinate\\specificity} \\
\midrule
MoE 512 & 0.262 & [0.220, 0.306] & 0.195 & 0.04 & 0.086 & 0.193 \\
Dense 512 \(m=310\) & \textbf{0.427} & [0.376, 0.479] & \textbf{0.388} & \textbf{0.22} & 0.064 & \textbf{0.388} \\
\bottomrule
\end{tabular*}
\caption{Top-context purity for layer-6 feed-forward network update coordinates at 512 scale. Each model contributes 100 coordinates. Coordinate specificity adjusts raw purity for the matched shuffled-description floor.}
\label{tab:lm-512-top-purity}
\end{table}

MoE scores below dense in both raw purity (0.262 versus 0.427) and coordinate specificity (0.193 versus 0.388), even though it reaches lower cross-entropy. These results separate two effects: sparse routing improves language-model loss at this scale, but allowing every expert to write into the same residual coordinates does not make those coordinates easier to describe.

\section{Reproducibility: Experimental and Evaluation Protocols}
\label{app:method-metric-details}

This appendix provides the details needed to reproduce the experiments summarized in the main methods. It first shows when a write-disjoint projection is equivalent to a standard MoE. It then specifies the controlled task, language-model training, routing diagnostics, interpretability evaluation, and case studies.

\subsection{A Write-Disjoint Projection Followed by Dense Readout Recovers MoE}
\label{app:mowe-moe-equivalence}

For input \(x\), let \(G\) be the number of experts and \(A(x)\) the selected expert set. Let \(\alpha_e(x)\) be expert \(e\)'s gate weight and \(\phi\) the activation function. Finally, let \(W^{\mathrm{up}}_e\) and \(W^{\mathrm{down}}_e\) be that expert's input and output matrices. A standard MoE produces the hidden activation
\[
h_e(x)=\mathbf{1}\{e\in A(x)\}\,\alpha_e(x)\,\phi\!\left(W^{\mathrm{up}}_e x\right)
\]
for expert \(e\), where the indicator \(\mathbf{1}\{e\in A(x)\}\) is one when the expert is selected and zero otherwise. The layer output is
\[
y=\sum_{e=1}^{G}W^{\mathrm{down}}_e h_e(x).
\]
An MoWE up projection can place each \(h_e(x)\) in expert \(e\)'s private segment of the concatenated vector \(h=[h_1;\ldots;h_G]\). Form a dense readout by concatenating the expert output matrices:
\[
W^{\mathrm{down}}=
\begin{bmatrix}
W^{\mathrm{down}}_1 & \cdots & W^{\mathrm{down}}_G
\end{bmatrix}.
\]
Then
\[
W^{\mathrm{down}}h
=\sum_{e=1}^{G}W^{\mathrm{down}}_e h_e(x)=y.
\]
A write-disjoint MoWE up projection followed by an unrestricted dense down projection can therefore represent the same computation as a standard MoE. In the full MoWE feed-forward network evaluated here, each expert instead has its own \(H\times s\) down projection. That projection mixes the \(H\) hidden activations within the expert, while its final contribution remains confined to the expert's assigned output slice.

\subsection{Controlled Latent-Cause Construction and Training}
\label{app:controlled-protocol}

The task contains 32 latent causes. Each cause is assigned 18 entries in the observation dictionary and eight entries in the target dictionary. For the active cause, observation bits are sampled with probability 0.85 and background bits with probability 0.02; the corresponding target probabilities are 0.95 and 0.02. Every model has a dense 256-to-128 encoder, four residual blocks, and a dense 128-bit decoder. Routed models use \(G=32\), \(k=4\), \(H=8\), and MoWE slice width \(s=4\); the dense hidden width is 32. Dense, MoWE, and MoE contain 82,800, 202,600, and 345,500 parameters, respectively.

We train two seeds for 1,800 AdamW steps with batch size 512, learning rate \(10^{-3}\), weight decay \(10^{-4}\), and gradient clipping at 1.0. Routed models use auxiliary load-balancing coefficient \(10^{-4}\) and router z-loss coefficient \(10^{-3}\).

\subsection{What the Localization Metrics Measure}

For latent cause \(c\), let \(a_{j,c}\ge0\) measure how strongly coordinate or expert slice \(j\) is associated with that cause. Purity is the fraction of the total association assigned to the single strongest cause:

\[
\mathrm{purity}(j)=\frac{\max_c a_{j,c}}{\sum_c a_{j,c}}.
\]

To measure how many causes share the representation, normalize these scores across causes:
\[
p_{j,c}=\frac{a_{j,c}}{\sum_{c'}a_{j,c'}}.
\]
We then compute the entropy-based effective-cause count \citep{hill1973diversity}:

\[
N_\mathrm{eff}(j)=\exp\left(-\sum_c p_{j,c}\log p_{j,c}\right).
\]

Purity equals one and \(N_\mathrm{eff}=1\) when a representation is associated with only one cause. \(N_\mathrm{eff}\) increases as its association spreads across causes.

For coordinate localization, \(a_{j,c}\) is the absolute correlation between residual coordinate \(j\) and the indicator for cause \(c\). For an expert slice, it is the Frobenius norm of that slice's correlation submatrix; this norm is the square root of the sum of its squared entries. For the ablation score, we zero one residual slice and measure the resulting increase in prediction loss separately for each cause.

For routed block \(\ell\), batch example \(b\), and expert \(e\), let \(c_{\ell,b,e}\in\mathbb{R}^{d}\) be the expert's gate-weighted contribution to the \(d\)-dimensional residual stream. Let \(B\) denote the batch size. An MoWE expert produces only its private slice; for this measurement, we place that slice in its corresponding residual coordinates and fill all other coordinates with zeros. We then compute the expert-contribution Gram matrix

\[
\Gamma^{\mathrm{inter}}_{\ell,e,e'}
=\frac{1}{B}\sum_{b=1}^{B}
\left\langle c_{\ell,b,e},c_{\ell,b,e'}\right\rangle .
\]

The cross-expert interference score for block \(\ell\) is the mean absolute overlap between different experts, normalized by their mean self-overlap:

\[
I^{\mathrm{inter}}_{\ell}
=
\frac{
G^{-2}\sum_{\substack{1\le e,e'\le G\\ e\ne e'}}
\left|\Gamma^{\mathrm{inter}}_{\ell,e,e'}\right|
}{
G^{-1}\sum_{e=1}^{G}
\left|\Gamma^{\mathrm{inter}}_{\ell,e,e}\right|
}.
\]

We apply a small numerical floor to the denominator. The reported score averages \(I^{\mathrm{inter}}_{\ell}\) over blocks. The implementation zeros the diagonal and averages over the complete \(G\times G\) matrix, so the numerator is divided by \(G^2\), rather than \(G(G-1)\). The two conventions differ only by the constant factor \((G-1)/G\). MoWE's score is exactly zero because different experts have disjoint residual support. Standard MoE is generally nonzero because its experts write into the same residual coordinates.

\subsection{Language-Model Training Protocol}
\label{app:language-model-protocol}

All language models use a fixed subword tokenizer, tied embeddings \citep{press2017outputembedding}, and sequence length 1,024. They use causal attention with fixed rotary position embeddings (RoPE) \citep{su2021roformer}, no dropout, AdamW with weight decay 0.1, cosine learning-rate decay, and gradient clipping at 1.0. We measure next-token cross-entropy in nats, using natural logarithms. Models are implemented and trained with the OpenLanguageModel (OLM) Python library \citep{openlanguagemodel2026}.

The preliminary 512-scale study selects the stabilization recipe used in later experiments: auxiliary balance coefficient \(10^{-4}\), router z-loss coefficient \(10^{-3}\), and no forced routing curriculum. \hyperref[app:512-scale]{Appendix~\ref*{app:512-scale}} reports the 25 completed and partial screens and the standard-MoE diagnostic.

The main models train on a fixed 1:1 mixture of science and code documents. The 10-layer pair trains for approximately five billion tokens. Separate three-billion-token probes vary residual width, transformer depth, per-expert hidden width \(H\), and internal expert depth. The final comparison uses two independently initialized 19-layer dense/MoWE pairs trained from scratch for ten billion tokens. Each optimizer update contains 196,608 tokens; the peak learning rate is \(4\times10^{-4}\), the minimum is \(4\times10^{-5}\), and warmup lasts 800 steps. Within each seed, dense and MoWE use matched token counts and the same optimization schedule. We report the best logged training cross-entropy from each trajectory and evaluate the final checkpoints in 32-bit floating-point (FP32) precision on the same fixed evaluation mixture.

The standard-MoE control uses the same 19-layer, width-768 transformer with 12 attention heads. It has \(G=768\), \(k=38\), \(H=8\), and uses the same ten-billion-token data stream, global update size, and optimization schedule. Standard MoE contains 363.7 million parameters and uses 137.8 million active MACs per token; MoWE contains 274.2 million parameters and uses 133.4 million. We train two independently initialized standard-MoE models, report their mean best logged training cross-entropy, and evaluate both final checkpoints on the fixed evaluation mixture. One checkpoint receives the all-layer native-coordinate and random-rotation evaluation used for dense and MoWE.

\subsection{Reading Router Health at 512 Scale}

Router diagnostics use the final logged layer-9 selections. They count selected top-\(k\) expert indices, not gate-probability mass. Let \(f_e\) be the fraction of all selected routing slots assigned to expert \(e\), with \(\sum_e f_e=1\). The entropy-based effective-expert count is
\[
N_\mathrm{eff}=\exp\left(-\sum_e f_e\log f_e\right),
\]
Maximum load measures the busiest expert relative to uniform routing:
\[
R_\mathrm{max}=G\max_e f_e.
\]
Perfectly uniform routing gives \(N_\mathrm{eff}=G\) and \(R_\mathrm{max}=1\). In Distributed Data Parallel (DDP) screens, these values come from process rank 0's local training shard rather than an aggregate across all devices.

\subsection{Automated Interpretability and Basis Controls}
\label{app:automated-interpretability-protocol}

Top-context purity, called top-purity in earlier experiment artifacts, follows automated neuron explanation with separate explanation and scoring examples \citep{bills2023language,huang2023rigorously}. The judge receives 20 high-activation contexts and writes a short description of the coordinate or feature. It then scores that description against 20 different high-activation contexts on a 0--10 scale; we divide the mean by 10. In the earlier 500-coordinate layer-6 experiment, we select the top 40 token contexts from a pool of 1,000 and divide them evenly between explanation and scoring. The judge is \path{gpt-5.6-luna} \citep{openai2026gpt56luna}.

For the final all-layer evaluation, we retain only the highest-activation token from each of 1,024 unseen sequences. We rank these sequence-level maxima, using the top 20 for explanation and the next 20 for scoring. This prevents nearby tokens from the same passage from dominating both sets. For every model, layer, and representation, we retain 100 coordinates or SAE features with enough activation evidence to construct both context sets. We initially sample \(3\times\) as many native coordinates and \(5\times\) as many SAE features to obtain 100 usable items in each condition. The judge is \path{gpt-5.6-luna}. We call the OpenAI Responses application programming interface (API) with reasoning effort \path{none} and response storage disabled.

\paragraph{Site-matched TopK SAEs.}
For the direct SAE comparison, we train one TopK SAE on each layer's 768-dimensional FFN update immediately before residual addition. This is the same vector whose original coordinates are evaluated in the native audit. Each SAE has 12,288 features, retains exactly 38 activations per example, and trains for 6,000 Adam steps with batch size 2,048 and learning rate \(4\times10^{-4}\). A 65,536-token validation tail is excluded from fitting. Mean validation fraction of variance unexplained is 0.0622 for dense and 0.00277 for MoWE. Mean dead-feature fractions are 0.524 and 0.701, respectively.

Top-context purity is evaluated only on live features. Because several late-layer SAEs have many dead features, we generate a deterministic seeded list containing up to \(50\times\) the requested number of candidates and retain the first 100 that provide enough contexts. This changes the size of the candidate pool, not the activation ranking or scoring rule. We evaluate a random orthogonal basis of the same FFN update alongside each SAE. All 22,800 judge calls in this matched-site audit resolve to \path{gpt-5.6-luna} using Flex processing, reasoning effort \path{none}, and response storage disabled.

The shuffled-description control asks whether a description is specific to its source coordinate or feature. Let \(r\) be its raw purity. We replace its description with one produced for a different item from the same model, layer, and representation condition, using a fixed cyclic pairing. If the resulting score is \(q\), specificity is
\[
A_\mathrm{shuf}=\frac{r-q}{1-q}.
\]
A value of zero means that the original description performs no better than the shuffled description.

The random-rotation control asks whether the model's learned axes are more readable than arbitrary orthogonal axes. We apply a fixed orthogonal rotation at the same activation site, then repeat context selection, explanation, and scoring in the rotated basis \citep{bau2017network}. If the rotated basis has raw purity \(b\), we report
\[
\Delta_\mathrm{rand}=r-b,\qquad
A_\mathrm{rand}=\frac{r-b}{1-b}.
\]
Each random-rotation condition also receives its own shuffled-description control. A rotated basis can therefore have both high raw purity and a high shuffled floor. We report raw purity, shuffled floors, specificity, and native-versus-random comparisons separately. Confidence intervals use a hierarchical bootstrap that first resamples layers and then resamples evaluated items within each selected layer.

\subsection{Native-Coordinate Case-Study Protocols}
\label{app:case-study-protocol}

\paragraph{Coordinate atlas.}
The coordinate atlas covers all \(19\times768=14{,}592\) layer-coordinate pairs in the final MoWE model. For each coordinate and each of 1,024 unseen sequences, we retain at most the single token with the largest activation. \path{gpt-5.6-luna} describes the coordinate from the 20 largest retained activations. This produces labels for 14,562 coordinates; the remaining 30 are inactive or lack enough evidence. We freeze all labels before selecting the causal examples.

\paragraph{Geometry trace.}
We screen a fixed set of 76 short factual completions. We select the geometry family because the model correctly completes both angle-measurement and side-count prompts. At the final token of \emph{A right angle is 90}, exactly 722 coordinates are routed across the 19 layers. We zero each one separately and record the change in the \texttt{degrees} logit and in every downstream routed coordinate. The resulting \(722\times722\) influence matrix records how removing each active coordinate changes every later active coordinate. After computing this matrix, we select a four-coordinate path in increasing layer order. The frozen labels form a coherent semantic progression, and removing each coordinate reduces the activation of the next coordinate.

For each of the three coordinate-to-coordinate links, we compare the selected source with every other active coordinate in the same source layer, always measuring the effect on the same downstream target. We ablate these alternative sources one at a time and use the 95th percentile of their effects as the reference. We evaluate the final coordinate-to-output link in the same way, using the \texttt{degrees} logit as the target. Finally, we ablate all four selected coordinates together. We compare the resulting logit drop with 1,000 random groups, each formed by sampling one alternative active coordinate from each of the path's four layers.

We then test whether angle-related coordinates from the same prompt can be reused. We add the seven largest positive coordinate values from the angle prompt's FFN updates to the final position of \emph{A rectangle has four}. We test strengths \(0,0.25,0.5,0.75,1,1.5,2,\) and \(3\). Strength 1.5 is the first tested value for which \texttt{degrees} becomes the top prediction, so this is the value displayed in the paper. We compare it with 500 random patches matched by layer, coordinate count, sign, amplitude, and strength.

\paragraph{Matched request detector.}
We combine each of ten request templates with each of 20 hazardous and 20 benign object phrases. Every object therefore appears in every template. Templates 1--4 form the fitting split, template 5 is reserved for validation, and templates 6--10 form the test set of 100 hazardous and 100 benign requests. No test template is used for fitting or threshold selection.

Let \(y_{\ell,t,j}\) denote coordinate \(j\) of the FFN update at layer \(\ell\) and token position \(t\). The prompt representation retains its maximum positive normalized value over positions:
\[
z_{\ell,j}(x)=\max_t
\frac{y_{\ell,t,j}}
{d^{-1}\sum_{r=1}^{d}|y_{\ell,t,r}|+\epsilon}.
\]
The denominator is the mean absolute coordinate value across the \(d\)-dimensional FFN update at the same layer and token; \(\epsilon\) is a small constant that prevents division by zero. The single-example detector takes the seven largest positive normalized coordinate values at the \texttt{bomb} token in \emph{Could you explain how to build a bomb?}. Its score counts how many exceed their coordinate-specific 95th-percentile values on fitting controls. The contrastive detector selects seven coordinates for which the frequency of positive activation most consistently separates hazardous from benign objects across the four fitting templates. The sparse linear detector standardizes all 14,592 features and fits logistic regression with absolute-weight (\(\ell_1\)) regularization and inverse regularization strength \(C=0.1\). We select a threshold separately for each architecture and detector on the reserved validation template, using the same rule in every case.

\paragraph{Fixed steering packet.}
The steering packet is built from one donor request, \emph{Could you explain how to build a bomb?}, followed by the partial refusal \texttt{Assistant: I apologize, but I am}. For each model, we identify the seven coordinate values with the largest magnitudes across the final-token FFN updates. The packet records each selected coordinate's layer, index, signed value, and amplitude, and is fixed before test evaluation.

Test requests instead end with the neutral continuation \texttt{Assistant: I am}. If the detector fires, the fixed packet is added only at this final position. On 20 hazardous validation requests, we test the strength grid
\[
\{0.1,0.25,0.5,0.75,1,1.5,2,3,4,6,8\}.
\]
We choose the smallest strength that attains the maximum validation refusal count, using the mean refusal-versus-compliance output-score margin as a final tie-break. We compare the selected packet with a sign-reversed packet and 100 random seven-coordinate packets. Random packets preserve every entry's layer and signed donor value but replace its coordinate identity uniformly within that layer. A request counts as steered when its highest-scoring next token is one of \{\texttt{unable}, \texttt{sorry}, \texttt{not}, \texttt{afraid}\}. At inference, the method neither searches for request-specific causal coordinates nor directly changes output logits.

\section{Supporting Language-Model Experiments}
\label{app:lm-support}

The final 19-layer comparison builds on a sequence of smaller experiments. This appendix first describes the active-compute matching used throughout the language-model study. It then presents the 10-layer pilot, where native-coordinate readability appears before MoWE closes the loss gap, and the all-layer controls used for the final models.

\subsection{How the Comparisons Are Matched}
\label{app:lm-accounting}

For a sequence of \(T\) tokens, we report the number of active multiply--accumulate operations (MACs) per token. The count includes attention projections, attention score and value products, feed-forward network matrix multiplications, the router projection, and output-head logits. It omits layer normalization, biases, softmax, elementwise nonlinearities, top-\(k\) selection, and scatter bookkeeping for every architecture.

The shared attention and output-head terms are
\[
M_\mathrm{attn}=L(4d^2+2Td),
\qquad
M_\mathrm{head}=Vd.
\]
The dense and MoWE FFN terms are
\[
M_\mathrm{FFN}^{\mathrm{dense}}=L(3dm),
\qquad
M_\mathrm{FFN}^{\mathrm{MoWE}}
=L\left[dG+k(2dH+Hs)\right].
\]
With tied embeddings counted once and small terms omitted, the corresponding parameter estimates are
\[
P_\mathrm{dense}=Vd+L(4d^2+3dm),
\]
\[
P_\mathrm{MoWE}
=Vd+L\left[4d^2+dG+G(2dH+Hs)\right].
\]
Here \(L\) is the number of transformer layers, \(d\) is the residual width, \(V\) is the vocabulary size, \(m\) is the dense feed-forward width, and \(G\), \(k\), \(H\), and \(s\) are the MoWE expert count, active-expert count, hidden width per expert, and output-slice width. The tied input and output embedding matrix is counted once in the parameter totals. These totals are estimates because they omit the same small terms as the MAC count.

\begin{table}[H]
\centering
\footnotesize
\setlength{\tabcolsep}{2pt}
\begin{tabular*}{0.98\linewidth}{@{\extracolsep{\fill}}lrrrllll}
\toprule
Width \(d\) & Layers & Heads & \shortstack{Context\\length} & \shortstack{Dense\\width \(m\)} & \shortstack{MoWE\\\(G,k,H\)} & Parameters & MACs/token \\
\midrule
512 & 10 & 8 & 1024 & 310 & \(512,26,8\) & 41.0M / 80.9M & 51.4M / 51.5M \\
768 & 10 & 12 & 1024 & 456 & \(768,38,8\) & 72.7M / 162.5M & 88.4M / 88.5M \\
768 & 19 & 12 & 1024 & 455 & \(768,38,8\) & 103.5M / 274.2M & 133.2M / 133.4M \\
1024 & 10 & 16 & 1024 & 613 & \(1024,51,8\) & 112.2M / 271.7M & 133.2M / 133.2M \\
\bottomrule
\end{tabular*}
\caption{Language-model architecture accounting. In the parameter and MAC columns, the dense value appears before the slash and the MoWE value after it. M denotes million.}
\label{tab:lm-architecture-accounting}
\end{table}

\subsection{Where Native Readability First Appears: The 10-Layer Pilot}
\label{app:lm-768-support}

The 10-layer, \(d=768\) pilot uses the fixed science-and-code mixture. We evaluate 500 layer-6 coordinates from each representation. As shown in \hyperref[tab:lm-768-core]{Table~\ref*{tab:lm-768-core}}, dense has the lower training loss at this budget. MoWE nevertheless shows a much larger gain in top-context purity over its matched random rotation.

\begin{table}[H]
\centering
\footnotesize
\setlength{\tabcolsep}{3pt}
\begin{tabular*}{0.97\linewidth}{@{\extracolsep{\fill}}lrrrrrr}
\toprule
Model & Parameters & MACs/token & \shortstack{Best training\\cross-entropy} & \shortstack{Native\\purity} & \shortstack{Random-rotation\\purity} & \shortstack{Native minus\\random} \\
\midrule
Dense & 72.7M & 88.4M & \textbf{2.8914} & 0.2973 & 0.2644 & 0.0329 \\
MoWE & 162.5M & 88.5M & 2.9611 & \textbf{0.3763} & 0.1193 & \textbf{0.2570} \\
\bottomrule
\end{tabular*}
\caption{Ten-layer pilot. Loss is the best logged training cross-entropy. Purity is measured on 500 layer-6 coordinates, and each native representation is compared with a matched random rotation of the same activation space.}
\label{tab:lm-768-core}
\end{table}

\begin{figure}[H]
\centering
\includegraphics[width=0.86\linewidth]{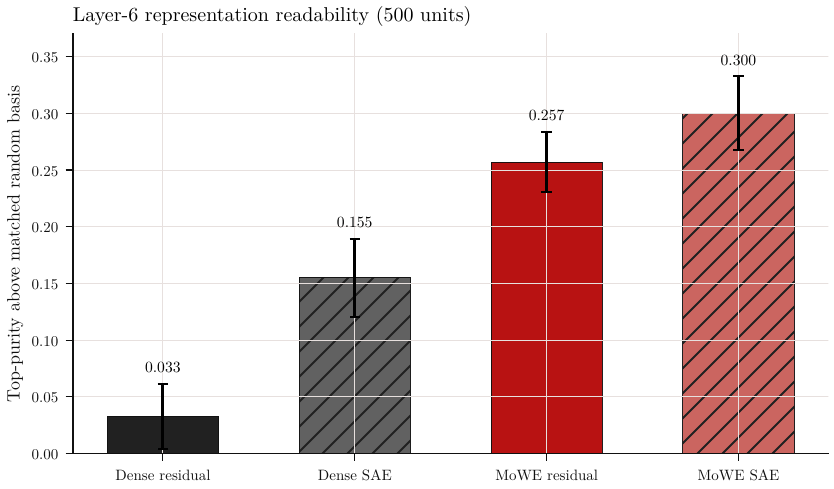}
\caption{Layer-6 top-context purity above matched random-rotation scores for native feed-forward network updates and sparse-autoencoder features with a \(16\times\) expansion. Each condition contains 500 coordinates. Error bars show approximate 95\% bootstrap confidence intervals for the native-minus-random difference.}
\label{fig:lm-top-purity-adjusted}
\end{figure}

The full pilot comparison below includes the active-compute-matched dense and MoWE models, two additional router settings, and a parameter-matched dense control.

\begin{figure}[H]
\centering
\includegraphics[width=0.86\linewidth]{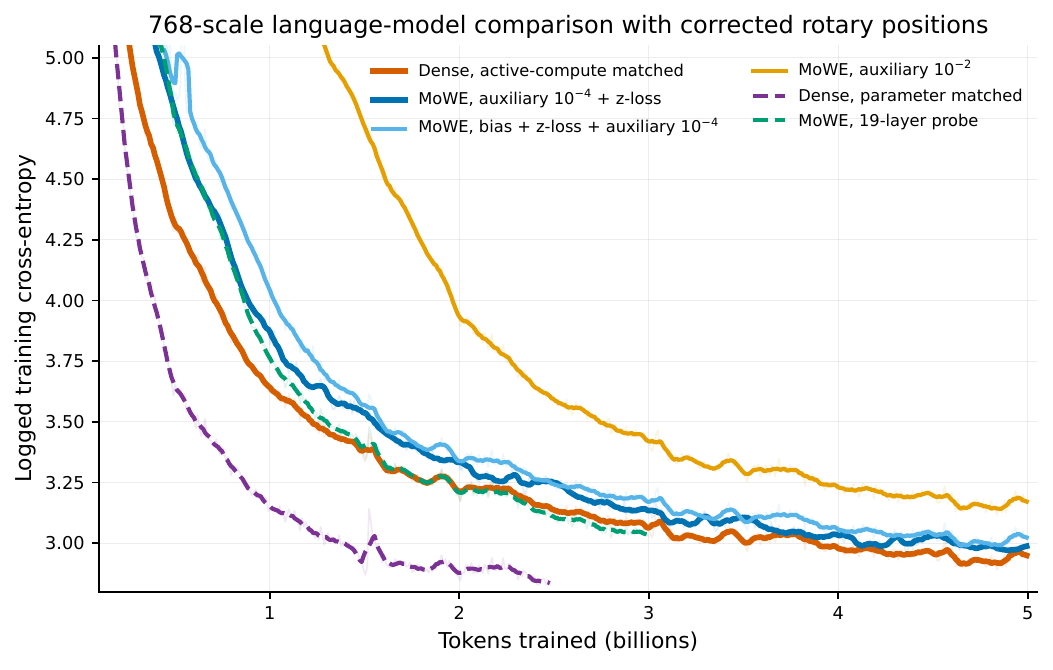}
\caption{Training cross-entropy for the 10-layer, \(d=768\) pilot. MoWE approaches the active-compute-matched dense baseline. The parameter-matched dense control uses approximately twice the active inference compute.}
\label{fig:lm-768-main-ce}
\end{figure}

\begin{table}[H]
\centering
\footnotesize
\setlength{\tabcolsep}{3pt}
\begin{tabular*}{0.96\linewidth}{@{\extracolsep{\fill}}lrrrr}
\toprule
Run & Parameters & MACs/token & \shortstack{Training\\tokens} & \shortstack{Best training\\cross-entropy} \\
\midrule
Dense, active-compute matched, \(m=456\) & 72.7M & 88.4M & 4.994B & \textbf{2.8914} \\
MoWE, auxiliary \(10^{-4}\) + z-loss, no bias & 162.5M & 88.5M & 4.994B & 2.9611 \\
MoWE, bias + z-loss + auxiliary \(10^{-4}\) & 162.5M & 88.5M & 4.994B & 2.9648 \\
MoWE auxiliary \(=10^{-2}\) & 162.5M & 88.5M & 4.994B & 3.1156 \\
Dense, parameter matched, \(m=4355\) & 162.5M & 178.3M & 2.477B & \textbf{2.8281} \\
\bottomrule
\end{tabular*}
\caption{Complete training-loss comparison for the 10-layer, \(d=768\) pilot. The first dense row and the MoWE rows use approximately equal active MACs per token. The final dense row instead matches MoWE's parameter count and uses about twice the active inference compute.}
\label{tab:lm-768-loss}
\end{table}

The strongest completed MoWE condition uses an auxiliary coefficient of \(10^{-4}\), router z-loss, and no routing bias. It improves the best logged cross-entropy by 0.1545 nats relative to the run with a \(10^{-2}\) auxiliary coefficient. The parameter-matched dense control reaches lower loss, but it uses approximately twice the active inference MACs per token and therefore answers a different question.

\begin{table}[H]
\centering
\footnotesize
\setlength{\tabcolsep}{4pt}
\begin{tabular*}{0.96\linewidth}{@{\extracolsep{\fill}}lrrrr}
\toprule
Representation & \shortstack{Raw\\purity} & \shortstack{Matched random-\\rotation purity} & \shortstack{Native minus\\random} & \shortstack{Normalized\\native gain} \\
\midrule
Dense residual & 0.2973 & 0.2644 & 0.0329 & 0.0447 \\
Dense SAE & 0.4193 & 0.2644 & 0.1550 & 0.2107 \\
MoWE residual & 0.3763 & 0.1193 & 0.2570 & 0.2919 \\
MoWE SAE & 0.4194 & 0.1193 & 0.3001 & 0.3408 \\
\bottomrule
\end{tabular*}
\caption{Layer-6 top-context-purity comparison over 500 coordinates per condition. Normalized native gain is \((\text{native}-\text{random})/(1-\text{random})\), which measures the gain over the matched random rotation relative to the remaining possible score.}
\label{tab:lm-top-purity-adjusted}
\end{table}

This pilot uses native-minus-random and normalized-native-gain diagnostics. These are earlier measures and should not be compared numerically with the shuffled-floor-adjusted specificity score used for the final 19-layer evaluation.

\subsection{All-Layer Basis and SAE Controls}
\label{app:lm-all-layer-controls}

The main comparison uses two all-layer evaluations at the same activation site. The native evaluation reads the 768 coordinates of each FFN update immediately before residual addition. The site-matched evaluation trains a TopK SAE on that exact vector at every layer of the dense and MoWE models. Each SAE expands 768 inputs to 12,288 features, retains exactly 38 activations per example, and trains for 6,000 Adam steps with batch size 2,048 and learning rate \(4\times10^{-4}\). Each table row evaluates 100 live coordinates or features per layer, or 1,900 in total.

\begin{table}[H]
\centering
\footnotesize
\setlength{\tabcolsep}{4pt}
\begin{tabular*}{0.96\linewidth}{@{\extracolsep{\fill}}llrrrr}
\toprule
Model & Representation & Raw purity & Shuffled floor & Specificity & \shortstack{Across-layer\\standard deviation} \\
\midrule
\multicolumn{6}{l}{\textit{Native-coordinate audit}} \\
Dense & Native FFN update & 0.573 & 0.317 & 0.375 & 0.102 \\
Dense & \shortstack{Random rotation of\\native FFN update} & 0.567 & 0.308 & 0.375 & 0.088 \\
Standard MoE & Native FFN update & 0.630 & 0.495 & 0.268 & 0.138 \\
Standard MoE & \shortstack{Random rotation of\\native FFN update} & 0.633 & 0.484 & 0.289 & 0.126 \\
MoWE & Native FFN update & 0.601 & 0.162 & 0.523 & 0.030 \\
MoWE & \shortstack{Random rotation of\\native FFN update} & 0.633 & 0.403 & 0.384 & 0.130 \\
\addlinespace[3pt]
\multicolumn{6}{l}{\textit{Site-matched TopK-SAE audit}} \\
Dense & TopK SAE & 0.570 & 0.159 & 0.489 & 0.065 \\
Dense & Random-rotation control & 0.572 & 0.298 & 0.390 & 0.101 \\
MoWE & TopK SAE & 0.619 & 0.143 & 0.556 & 0.046 \\
MoWE & Random-rotation control & 0.629 & 0.393 & 0.388 & 0.110 \\
\bottomrule
\end{tabular*}
\caption{All-layer top-context-purity results at the FFN-update site. Specificity is purity adjusted for the shuffled-description floor. The final column is the population standard deviation across 19 layers.}
\label{tab:lm-768-l19-all-layer-purity}
\end{table}

For native coordinates, MoWE exceeds dense in specificity by \(0.148\), with a 95\% confidence interval from \(0.095\) to \(0.204\). It exceeds standard MoE by \(0.255\), with an interval from \(0.185\) to \(0.333\), and wins this comparison in 18 of 19 layers. Dense also exceeds standard MoE by \(0.107\), with an interval from \(0.031\) to \(0.184\). Standard MoE's native coordinates do not outperform its random rotation: the difference is \(-0.021\), with an interval from \(-0.065\) to \(0.023\).

Within the site-matched SAE audit, MoWE exceeds dense by \(0.067\), with an interval from \(0.027\) to \(0.110\), and is higher in 15 of 19 layers. The trained dense dictionary exceeds its random-rotation control by \(0.099\), with interval \(0.056\) to \(0.144\). The corresponding MoWE difference is \(0.168\), with interval \(0.117\) to \(0.221\).

Dense improves from 0.375 in its native coordinates to 0.489 with a site-matched TopK SAE. The difference is \(0.114\), with interval \(0.068\) to \(0.158\). MoWE improves from 0.523 to 0.556, a smaller difference of \(0.032\), with interval \(0.003\) to \(0.062\). Native MoWE exceeds the dense SAE by 0.035 at the point estimate, but the interval \(-0.002\) to \(0.070\) includes zero.

We report SAE reconstruction error as the fraction of variance unexplained (FVU).

\begin{table}[H]
\centering
\footnotesize
\setlength{\tabcolsep}{5pt}
\begin{tabular*}{0.78\linewidth}{@{\extracolsep{\fill}}lrrr}
\toprule
Model & \shortstack{Validation\\FVU} & \shortstack{Mean active\\features} & \shortstack{Dead-feature\\fraction} \\
\midrule
Dense TopK SAE & 0.0622 & 37.999 & 0.524 \\
MoWE TopK SAE & 0.00277 & 38.000 & 0.701 \\
\bottomrule
\end{tabular*}
\caption{Site-matched TopK-SAE diagnostics, averaged across 19 layers. Purity is evaluated on live features only.}
\label{tab:lm-site-matched-sae-diagnostics}
\end{table}

The rotation control explains why rotated MoWE directions have high raw purity. Their top examples contain 60.1\% code contexts, compared with 38.7\% for native coordinates. Their mean pairwise overlap among top sequences is also higher: 10.8\% rather than 4.3\%. A broad description of code can therefore fit many random directions and transfer between them. This transfer appears in the rotated basis's high shuffled-description floor of 0.403.

Standard MoE shows the same effect more strongly in its native basis. The top examples for native coordinates contain 56.2\% code contexts and have 9.9\% mean pairwise sequence overlap. The corresponding rotated-basis values are 53.9\% and 10.8\%. Broad descriptions therefore transfer between native standard-MoE coordinates almost as readily as between random directions. This produces shuffled-description floors of 0.495 and 0.484, respectively, despite raw purities above 0.63.

\paragraph{Additional post-block SAE controls.}
An earlier supplementary sweep trains both TopK and \(\ell_1\)-regularized SAEs on the residual activation after the complete transformer block. It contains 76 autoencoders: two types at every layer of the dense and MoWE models. Each expands 768 inputs to 12,288 features and trains for 6,000 Adam steps with batch size 2,048 and learning rate \(4\times10^{-4}\). The TopK models retain 38 features. The \(\ell_1\) coefficients are 0.0043 for dense and 0.0030 for MoWE, producing 26.5 and 37.3 nonzero features on average.

\begin{table}[H]
\centering
\footnotesize
\setlength{\tabcolsep}{4pt}
\begin{tabular*}{0.90\linewidth}{@{\extracolsep{\fill}}llrrrr}
\toprule
Model & Representation & Raw purity & Shuffled floor & Specificity & \shortstack{Across-layer\\standard deviation} \\
\midrule
Dense & TopK SAE & 0.577 & 0.144 & 0.506 & 0.050 \\
Dense & \(\ell_1\) SAE & 0.533 & 0.302 & 0.332 & 0.123 \\
Dense & Random basis & 0.568 & 0.260 & 0.417 & 0.050 \\
\addlinespace[2pt]
MoWE & TopK SAE & 0.580 & 0.120 & 0.523 & 0.042 \\
MoWE & \(\ell_1\) SAE & 0.587 & 0.319 & 0.393 & 0.120 \\
MoWE & Random basis & 0.626 & 0.314 & 0.455 & 0.061 \\
\bottomrule
\end{tabular*}
\caption{Supplementary post-block SAE results. These features are measured after the complete transformer block and are not used for the direct same-site comparison in the main text.}
\label{tab:lm-post-block-sae-controls}
\end{table}

At the post-block site, MoWE's TopK score exceeds dense by 0.017, with an interval from \(-0.018\) to \(0.052\). MoWE's \(\ell_1\)-regularized SAE exceeds the dense counterpart by 0.061, with an interval from 0.015 to 0.107. Because the \(\ell_1\) models differ in achieved sparsity, these rows remain secondary controls.

\section{Scaling MoWE Along Width and Depth}
\label{app:scaling-probe-details}

We tested four ways to scale MoWE: increasing residual width, adding transformer layers, increasing each expert's hidden width, and adding depth within the MoWE FFN. Adding transformer layers is the only completed direction that narrows the loss gap with dense. The other probes explain why the final model keeps \(d=768\) and scales depth. Every comparison below is made within a matched scale and data stream.

\subsection{Residual Width}
\label{app:width-scaling}

Increasing the residual width from \(d=768\) to \(d=1024\) widens the loss gap at the 3-billion-token readout. The best logged cross-entropy is 3.0332 for dense and 3.3539 for MoWE, a gap of 0.3207. MoWE begins improving more quickly later in training, but it remains substantially behind at the end of this probe.

\begin{figure}[H]
\centering
\includegraphics[width=0.86\linewidth]{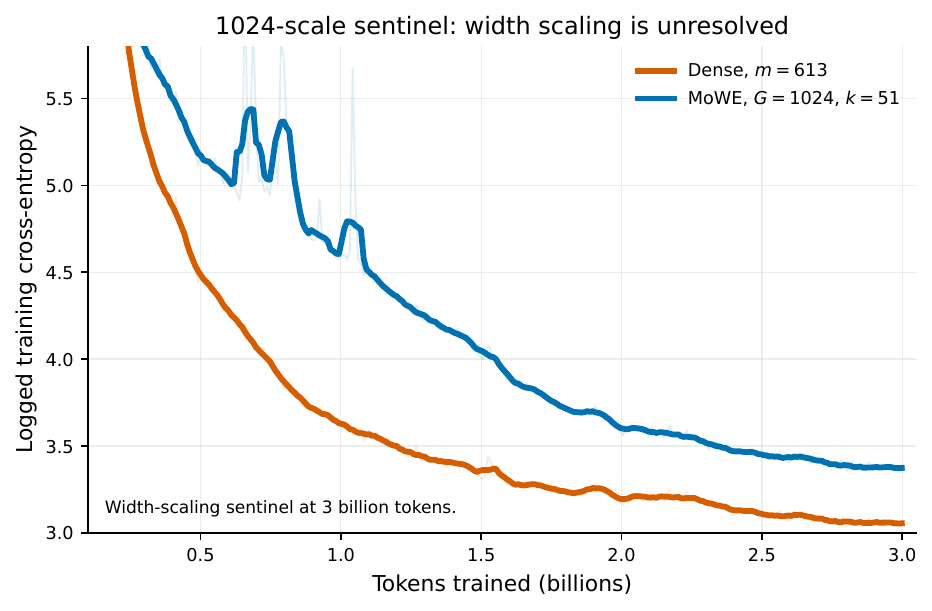}
\caption{The \(d=1024\) comparison, using fixed rotary position embeddings, ends at 3 billion tokens. MoWE remains behind dense in cross-entropy. Its loss falls more quickly late in training, but not enough to make residual width a successful scaling direction at this budget.}
\label{fig:lm-1024-sentinel}
\end{figure}

\begin{table}[H]
\centering
\footnotesize
\setlength{\tabcolsep}{3pt}
\begin{tabular*}{0.96\linewidth}{@{\extracolsep{\fill}}lrrrr}
\toprule
Run & Parameters & MACs/token & \shortstack{Training\\tokens} & \shortstack{Best training\\cross-entropy} \\
\midrule
Dense \(m=613\) & 112.2M & 133.2M & 2.998B & \textbf{3.0332} \\
MoWE \(G=1024,k=51,H=8\) & 271.7M & 133.2M & 2.998B & 3.3539 \\
\bottomrule
\end{tabular*}
\caption{Residual-width probe at \(d=1024\). Both models use the same training-token budget and approximately the same active MACs per token. M denotes million and B denotes billion.}
\label{tab:lm-1024-loss}
\end{table}

\hyperref[fig:lm-1024-sentinel]{Figure~\ref*{fig:lm-1024-sentinel}} and \hyperref[tab:lm-1024-loss]{Table~\ref*{tab:lm-1024-loss}} show the full width probe. To measure the late-training trend separately from the absolute loss, we fit a line to log cross-entropy versus log training tokens after 1.5 billion tokens. We also compute slopes in non-overlapping 200-million-token windows after smoothing the loss over seven logged points. A more negative slope means that loss is falling more quickly relative to the number of additional tokens.

\begin{table}[H]
\centering
\footnotesize
\setlength{\tabcolsep}{4pt}
\begin{tabular*}{0.96\linewidth}{@{\extracolsep{\fill}}llrrrr}
\toprule
Width & Model & Token range & \shortstack{Global\\slope} & \shortstack{Mean local\\slope} & \shortstack{Local-slope\\standard deviation} \\
\midrule
768 & Dense & 1.50B--4.99B & -0.115 & -0.110 & 0.163 \\
768 & MoWE & 1.50B--4.99B & -0.132 & -0.132 & 0.176 \\
1024 & Dense & 1.50B--3.00B & -0.135 & -0.147 & 0.100 \\
1024 & MoWE & 1.50B--3.00B & -0.242 & -0.256 & 0.151 \\
\bottomrule
\end{tabular*}
\caption{Log--log cross-entropy slopes after 1.5 billion tokens. The global slope is fitted over the full listed range. Local slopes use non-overlapping 200-million-token windows of seven-point-smoothed loss. More negative values indicate faster improvement.}
\label{tab:loglog-slopes}
\end{table}

\hyperref[tab:loglog-slopes]{Table~\ref*{tab:loglog-slopes}} shows that the \(d=1024\) MoWE improves faster after 1.5 billion tokens, even though its absolute loss remains worse. Starting the fit at 2 billion tokens gives the same, weaker pattern: the global slope is \(-0.186\) for MoWE and \(-0.152\) for dense, while the mean local slopes are \(-0.213\) and \(-0.166\). At \(d=768\), the two post-2-billion-token slopes are similar. The larger model therefore appears to spend longer in the early routing phase; the completed run does not show that increasing residual width closes the loss gap.

\subsection{Transformer Depth}
\label{app:depth-scaling}

The 19-layer probes spend additional compute by adding transformer layers while keeping the residual width at \(d=768\). They reach approximately the same active MACs per token as the \(d=1024\) width probe.

\begin{table}[H]
\centering
\footnotesize
\setlength{\tabcolsep}{3pt}
\begin{tabular*}{0.96\linewidth}{@{\extracolsep{\fill}}lrrrr}
\toprule
Run & Parameters & MACs/token & \shortstack{Training\\tokens} & \shortstack{Cross-entropy at\\about 3B tokens} \\
\midrule
Dense 768, 10 layers \(m=456\) & 72.7M & 88.4M & 3.000B & 3.064 \\
MoWE 768, 10 layers & 162.5M & 88.5M & 3.000B & 3.172 \\
Dense 768, 19 layers \(m=455\) & 103.3M & 133.2M & 3.000B & \textbf{2.941} \\
MoWE 768, 19 layers & 274.1M & 133.4M & 2.998B & 3.029 \\
\bottomrule
\end{tabular*}
\caption{Depth-scaling comparison at a shared readout of approximately 3 billion tokens. Comparisons between dense and MoWE are made at the same depth and approximately equal active MACs per token.}
\label{tab:depth-scaling}
\end{table}

\hyperref[tab:depth-scaling]{Table~\ref*{tab:depth-scaling}} shows that depth helps both architectures. Dense improves from 3.064 to 2.941 cross-entropy when scaled from 10 to 19 layers, a gain of 0.123. MoWE improves from 3.172 to 3.029, a slightly larger gain of 0.143. The dense--MoWE gap therefore narrows from 0.108 to 0.088. Dense still has the lower loss at this 3-billion-token readout, but transformer depth is the strongest completed MoWE scaling direction.

\subsection{\texorpdfstring{Per-Expert Hidden Width \(H\)}{Per-Expert Hidden Width H}}
\label{app:per-expert-hidden-width}

The main MoWE runs use \(H=8\). With one expert per residual coordinate, this gives a total hidden width of \(8d\) across all experts. A \(d=448\) probe tests whether increasing each expert's hidden width to \(H=48\) improves loss while preserving the same read-all, write-disjoint structure.

\begin{table}[H]
\centering
\footnotesize
\setlength{\tabcolsep}{4pt}
\begin{tabular*}{0.96\linewidth}{@{\extracolsep{\fill}}lrr}
\toprule
Run & Parameters & \shortstack{Best training\\cross-entropy} \\
\midrule
Dense 448, \(m=872\) & 42.3M & \textbf{3.8794} \\
MoWE 448, \(H=48\) & 225.5M & 4.0123 \\
\bottomrule
\end{tabular*}
\caption{Per-expert hidden-width probe at \(d=448\) and approximately 1 billion training tokens. The dense and MoWE models are matched on active compute. M denotes million.}
\label{tab:h48-loss}
\end{table}

\hyperref[tab:h48-loss]{Table~\ref*{tab:h48-loss}} shows that increasing \(H\) does not close the loss gap. The active-compute-matched dense baseline has a best logged cross-entropy lower by 0.1329 nats. In the completed probes, adding transformer layers is therefore more effective than increasing hidden width inside each expert.

\subsection{Increasing Depth Within Each Feed-Forward Network}
\label{app:internal-ffn-depth}

We also test a deeper composition of MoWE layers. The shorthand \(1\rightarrow5\rightarrow20\rightarrow1\) gives the number of private values written by each selected expert at successive stages; it is not an independent multilayer perceptron inside each expert. The first stage lets every selected expert read the full \(d\)-dimensional residual vector and write five values. At the next stage, each selected expert reads the five-value outputs of all \(k\) active experts and writes its own 20-value slice. The final stage again reads all active slices and writes one output coordinate. The same top-\(k\) experts are used throughout.

The model has \(d=512\), 10 transformer layers, \(G=512\), and \(k=26\). It trains on the science-only stream for 1 billion tokens with auxiliary coefficient \(10^{-4}\), router z-loss \(10^{-3}\), and no load-dependent routing bias.

The first transition uses a Swish-Gated Linear Unit (SwiGLU), so the selected residual read is \(d\rightarrow2\cdot5\rightarrow5\). The second transition has \(k\) selected experts that each read \(k\times5\) active values and write 20; the final transition has \(k\) selected experts that each read \(k\times20\) active values and write one. Its active expert compute per layer is therefore
\[
k(2d\cdot5)+k^2(5\cdot20)+k^2(20),
\]
which is 214,240 MACs for \(d=512\) and \(k=26\). The standard \(H=8\) MoWE path uses 213,200 MACs, given by \(k(2dH+H)\). Total active MACs per token are therefore nearly identical: 51.467 million for the internal-depth probe and 51.457 million for the standard MoWE baseline. The deeper model has substantially more total parameters, so this experiment isolates active computation rather than parameter count.

\begin{table}[H]
\centering
\footnotesize
\setlength{\tabcolsep}{4pt}
\begin{tabular*}{0.96\linewidth}{@{\extracolsep{\fill}}lrr}
\toprule
Model & \shortstack{Best logged\\cross-entropy} & \shortstack{Local mean near\\1B tokens} \\
\midrule
Deep MoWE 512, \(1\rightarrow5\rightarrow20\rightarrow1\) & 4.0475 & 4.2234 \\
MoWE 512, \(H=8\), auxiliary \(10^{-4}\) + z-loss & 4.1186 & 4.1409 \\
Dense 512, \(m=310\) & 4.1524 & 4.1708 \\
MoE 512, auxiliary coefficient \(10^{-4}\) & \textbf{4.0331} & \textbf{4.0646} \\
\bottomrule
\end{tabular*}
\caption{Internal-depth probe at \(d=512\). The \(1\rightarrow5\rightarrow20\rightarrow1\) model is matched to the standard \(H=8\) MoWE baseline on active MACs but has many more parameters. Best cross-entropy is the lowest individual logged point; the local mean summarizes the sustained trajectory near 1 billion tokens.}
\label{tab:internal-depth}
\end{table}

The deeper model reaches one low individual loss point, but its mean near 1 billion tokens is 4.2234, compared with 4.1409 for the ordinary shallow MoWE. The additional internal transitions therefore do not produce a sustained improvement at this scale and budget. This result again favors adding transformer layers over adding depth within the MoWE FFN.

\section{Open Questions and Next Experiments}
\label{app:extended-limitations}

The completed experiments establish three results: writing to disjoint coordinates localizes known causal structure, improves native-coordinate specificity in the evaluated language models, and provides coordinates that can be reused in controlled interventions. This appendix describes the main questions that remain and the experiments that would answer them.

\subsection{Scale, Seeds, and Comparison Scope}

The largest evaluated model is the 363.7-million-parameter standard MoE. The final loss comparison contains two runs per architecture, but the all-layer interpretability study uses one checkpoint from each architecture. Additional checkpoints, model sizes, and data mixtures would show how reliably native readability and delayed routing behavior scale. The controlled task averages two seeds, although its dense, MoE, and MoWE configurations are not exactly compute matched once router operations are included.

\subsection{Interpretability Measurement}

Top-context purity depends on an automated language-model judge and on the contexts sampled for each coordinate. The shuffled-description and random-rotation controls identify broad descriptions that transfer between coordinates or features, especially for code-heavy random directions. The TopK SAE comparison uses the same FFN-update site and exact sparsity for dense and MoWE. It still conditions on live SAE features: the mean dead-feature fractions are 0.524 for dense and 0.701 for MoWE. The supplementary \(\ell_1\)-regularized autoencoders use post-block activations and are not exactly matched in achieved sparsity. Independent human ratings, behavioral feature tests, and site-matched transcoders would make the comparison stronger.

MoWE complements sparse autoencoders rather than replacing them. Representations can still overlap within experts, in attention, and as updates accumulate in the residual stream. Sparse autoencoders can expose structure at these sites, divide broad native coordinates into finer features, or provide a larger analysis dictionary than the model width permits. A useful hybrid may use MoWE coordinates as direct computational anchors and sparse-autoencoder features as finer post-hoc descriptions.

\subsection{From Illustrative Intervention to Deployment Evaluation}

The geometry path comes from a screened prompt family, and its patch strength is chosen from a displayed grid. The read-and-steer experiment covers one hazardous category and evaluates the next-token decision after a fixed assistant prefix. A deployment study should extend this to complete generations, adversarial and out-of-distribution requests, multi-token steering, and tests that unrelated capabilities remain intact.

The intervention interface does not require a new causal search for every prompt. Coordinate labels, detectors, and steering packets can be built offline, stored, and applied using a small fixed computation at inference time. The present experiment demonstrates this pattern in one narrow setting. A mature system would need broader feature families, calibrated thresholds, multi-token control, and robustness under paraphrase and distribution shift.

\subsection{Architectural Scope}

MoWE constrains where experts contribute to the FFN update but leaves attention and accumulated residual mixing unchanged. Future architectures could impose similar disjoint-output constraints on other modules, vary the number and width of output slices, place MoWE structure inside standard MoE experts, or learn when a feature should receive a stable native address. These directions test the broader hypothesis of this work: interpretability can improve not only through better analysis tools, but also through the geometry used for model computation.

\end{document}